\documentclass[11pt]{article}

\usepackage{amsmath,amsfonts,bm}

\def\eqref#1{equation~\ref{#1}}

\def\1{\bm{1}}

\def\rmS{{\mathbf{S}}}

\def\vmu{{\bm{\mu}}}

\def\vd{{\bm{d}}}

\def\vy{{\bm{y}}}

\def\mA{{\bm{A}}}
\def\mB{{\bm{B}}}

\def\mX{{\bm{X}}}

\DeclareMathAlphabet{\mathsfit}{\encodingdefault}{\sfdefault}{m}{sl}
\SetMathAlphabet{\mathsfit}{bold}{\encodingdefault}{\sfdefault}{bx}{n}

\def\gG{{\mathcal{G}}}

\def\sN{{\mathbb{N}}}

\def\sR{{\mathbb{R}}}

\usepackage{hyperref}
\usepackage{comment}
\usepackage{url}

\usepackage{algorithm}
\usepackage{algpseudocode}
\usepackage{multirow}
\usepackage{booktabs}

\usepackage{graphicx}
\usepackage{enumitem,amssymb}

\usepackage{tikz}
\usetikzlibrary{matrix,arrows.meta,positioning,calc,decorations.pathreplacing}
\usepackage{subcaption}
\usepackage{todonotes}
\usepackage{natbib}

\usepackage{authblk}

\algrenewcommand\algorithmicrequire{\textbf{Input:}}
\algrenewcommand\algorithmicensure{\textbf{Output:}}

\newlist{todolist}{itemize}{2}
\setlist[todolist]{label=$\square$}
\usepackage{pifont}

\title{FUSE: Fast Semi-Supervised Node Embedding Learning via Structural and Label-Aware Optimization}

\author[1]{Sujan Chakraborty\thanks{These authors contributed equally.}}
\author[2]{Rahul Bordoloi$^*$}
\author[4]{Anindya Sengupta}
\author[2,3]{Olaf Wolkenhauer}
\author[1]{Saptarshi Bej\thanks{Corresponding author: \texttt{sbej7042@iisertvm.ac.in}}}

\affil[1]{School of Data Science, IISER Thiruvananthapuram, India}
\affil[2]{Institute of Computer Science, University of Rostock, Germany}
\affil[3]{Leibniz-Institute for Food Systems Biology; Technical University of Munich, Freising; Germany}
\affil[4]{Texas A\&M University, USA}

\newcommand{\arxivlink}{\href{https://ogb.stanford.edu/docs/nodeprop/\#ogbn-arxiv}{ArXiV}}
\begin{document}

\maketitle
\begin{abstract} 
Graph-based learning is a cornerstone for analyzing structured data, with node classification as a central task. However, in many real-world graphs, nodes lack informative feature vectors, leaving only neighborhood connectivity and class labels as available signals. In such cases, effective classification hinges on learning node embeddings that capture structural roles and topological context. We introduce a fast semi-supervised embedding framework that jointly optimizes three complementary objectives: (i) unsupervised structure preservation via scalable modularity approximation, (ii) supervised regularization to minimize intra-class variance among labeled nodes, and (iii) semi-supervised propagation that refines unlabeled nodes through random-walk-based label spreading with attention-weighted similarity. These components are unified into a single iterative optimization scheme, yielding high-quality node embeddings. On standard benchmarks, our method consistently achieves classification accuracy at par with or superior to state-of-the-art approaches, while requiring significantly less computational cost.
\end{abstract}

\section{Introduction} \label{sec:intro}

Graph-based learning has emerged as a powerful paradigm for analyzing structured data, with applications in social networks \citep{social_networks_gnn}, citation graphs \citep{luo2023citationsum}, knowledge graphs \citep{knwoledge_graphs_gnn}, and recommendation systems \citep{industrial_applications_gnn, recommender_systems_gnn}. A central task is node classification, where a subset of nodes are labeled and the goal is to predict the labels of the remaining ones \citep{luo2024hybridcurriculum}. This task is typically facilitated by node embeddings \( \mX \in \sR^{|V|\times k} \) that capture graph structure \citep{Xiao_Wang_Dai_Guo_2021}.  

In practice, node embeddings may not be explicitly available, especially in newly constructed or rapidly evolving graphs, even when partial labels are known. Existing approaches often rely on unsupervised \citep{deepmincut} or self-supervised \citep{dgi} embedding generation, or directly employ Graph Neural Networks (GNNs) such as GCN \citep{GCN}, GAT \citep{GAT}, and GraphSAGE \citep{GraphSAGE} in a semi-supervised fashion. In addition, there are a few semi-supervised approaches that combine GNNs as encoders and customized classifiers to solve node classification problems  \citep{lee2022grafn, yan2023revar}. The given features are enhanced using these semi-supervised node representation algorithms. However, when embeddings are missing, initializing GNNs with random embeddings is ineffective for downstream tasks. A more efficient strategy is to generate structured initial embeddings via unsupervised or self-supervised approaches, and then refine them with GNNs \citep{GraphSAGE, pretraining_GNN}.  

We propose a fast semi-supervised embedding generation framework designed specifically for cases where node embeddings are unavailable. Our method integrates three complementary optimization components:  
\begin{enumerate}
    \item \textbf{Unsupervised structure preservation}, capturing global connectivity through a novel scalable approximation of graph modularity \citep{Modularity, linear_modularity_yazdanparast}.  
    \item \textbf{Supervised regularization}, aligning labeled nodes within the same class via compactness constraints. 
    \item \textbf{Semi-supervised propagation}, refining unlabeled nodes using random-walk-based label propagation~\citep{label_propagation} combined with attention-driven similarity weighting~\citep{node_attention}.  
\end{enumerate}

By unifying these three components into a single iterative gradient ascent framework, our approach produces high-quality node embeddings quickly and without requiring pre-existing features. The fast convergence of the optimization procedure can make it well-suited to settings where labels are introduced incrementally, making it especially relevant in real-world applications such as recommendation \citep{pei2020geom}, cybersecurity \citep{fang2022lmtracker}, and financial transaction monitoring \citep{bukhori2023inductive}, where embeddings must be updated on the fly.  

We evaluate our approach on standard benchmarks including Cora \citep{cora}, Citeseer \citep{citeseer}, WikiCS \citep{wikics}, Amazon Photo \citep{photo}, PubMed \citep{pubmed} and ArXiV \citep{hu2020open}. We compare against widely used unsupervised methods such as Node2Vec \citep{node2vec}, DeepWalk \citep{deepwalk}, VGAE \citep{vgae}, M-NMF \citep{wang2017m}, the self-supervised DGI \citep{dgi}, two semi supervised baselines GraFN \citep{lee2022grafn}, ReVAR \citep{yan2023revar} and precomputed embeddings. Downstream classification performance is assessed using GCN \citep{GCN}, GAT \citep{GAT}, and GraphSAGE \citep{GraphSAGE}.  

\textbf{Contributions.} Our main contributions are as follows:
\begin{enumerate}
    \item We introduce a fast semi-supervised embedding generation algorithm that requires no predefined node embeddings. 
    \item In particular, we propose a linear time approximation of the graph modularity and its gradient (which avoids costly eigen-decompositions), which is fundamental to our fast embedding generation process.
    \item Notably, the algorithm uses labels if available, but can be adapted to scenarios where labels are completely unavailable with some compromise in performance.
    \item We design a unified optimization framework that equally integrates unsupervised, supervised, and semi-supervised components.    
\end{enumerate}

\section{Related Work} \label{sec:related}

Our approach connects to several lines of research: unsupervised embedding methods, self-supervised, semi-supervised embedding methods, graph neural network baselines, and modularity-driven optimization.  

\textbf{Unsupervised node embedding.} Random-walk-based approaches such as DeepWalk \citep{deepwalk} and Node2Vec \citep{node2vec} learn node representations by applying Skip-Gram training to sequences generated from biased or unbiased random walks. Variational Graph Auto-Encoders (VGAE) \citep{vgae} extend autoencoding ideas to graphs by using a GCN encoder with a latent Gaussian distribution, achieving strong results in unsupervised link prediction.  Another method, M-NMF \cite{wang2017m} learn node embeddings by factorizing the graph structure without using any label information. It integrates both the network’s local structure (e.g., adjacency information) and global community structure (e.g., modularity) into a joint factorization framework. These methods demonstrate that structural information alone can be leveraged to build embeddings, since they are agnostic to label information.

\textbf{Self-supervised learning.} Contrastive frameworks such as Deep Graph Infomax (DGI) \citep{dgi} maximize mutual information between local node embeddings and global summaries, enabling representation learning without labels. Other approaches (e.g., SL-GAT \citep{node_attention}) refine attention-based architectures with self-supervised objectives. These methods reduce the reliance on labeled data but typically incur significant computational overhead.  

\textbf{Semi-supervised learning.} Semi-supervised methods like GraFN \citep{lee2022grafn} and ReVAR \citep{yan2023revar} address the limitations of purely supervised or self-supervised graph learning by combining a small amount of labeled data with structural information. GraFN aligns class predictions across augmented graph views to improve class-discriminative representations by combining self-supervised and label-guided methods, while ReVAR, which is specifically designed for imbalanced scenarios, introduces variance-based regularization to mitigate class imbalance.

\textbf{Graph neural networks for classification.} Semi-supervised GNNs such as GCN \citep{GCN}, GAT \citep{GAT}, and GraphSAGE \citep{GraphSAGE} refine embeddings through message passing and neighborhood aggregation, making them effective classifiers once initial embeddings are provided. Recent surveys highlight their utility across domains including social networks \citep{social_networks_gnn}, knowledge graphs \citep{knwoledge_graphs_gnn}, and recommender systems \citep{industrial_applications_gnn, recommender_systems_gnn}. However, initializing GNNs with random embeddings is ineffective for downstream tasks \citep{embedding_learning}, motivating the need for fast strategies that generate embeddings from scratch. It is to be noted that throughout the tables provided, we used ``SAGE'' to represent GraphSAGE primarily due to space constraints. 

\textbf{Connections of proposed objective to prior works.}  
Our proposed objective unifies three complementary components, each drawing inspiration from existing lines of research:  
\begin{enumerate}
    \item \textbf{Unsupervised structural component.} Modularity \citep{Modularity} and its scalable variants \citep{linear_modularity_yazdanparast, lu2018adaptive} have long been used for identifying communities in graphs. Neural formulations such as DGCLUSTER \citep{dgcluster} further relaxed modularity maximization into differentiable objectives. Inspired by this line of work, we design an unsupervised objective that preserves structural regularities while avoiding the computational overhead of spectral methods.  
    \item \textbf{Supervised label-aware component.} Semi-supervised GNNs such as GCN, GAT, and GraphSAGE \citep{GCN, GAT, GraphSAGE} incorporate label signals during message passing to improve classification performance. We adapt this idea directly at the embedding generation stage, encouraging nodes with the same label to have structurally similar embeddings. This distinguishes our approach from prior GNN methods, which rely on node features.  
    \item \textbf{Semi-supervised propagation component.} Label propagation \citep{label_propagation} and attention-based refinements such as SL-GAT \citep{node_attention} have demonstrated the ability to diffuse label information across the graph in a scalable way. We build on these insights by incorporating a random-walk-based propagation mechanism that guides the embeddings of unlabeled nodes toward those of reachable labeled nodes.  
\end{enumerate}

This work bridges these strands by proposing a fast semi-supervised algorithm that avoids dependence on node features, while combining the strengths of unsupervised structural preservation, supervised label regularization, and semi-supervised propagation.

\section{The FUSE Algorithm}\label{sec:method}
Our approach, Fast Unified Semi-supervised Node Embedding Learning from Scratch (FUSE) combines linearized modularity optimization with supervised regularization and semi-supervised label propagation to generate embeddings that are both structurally coherent and class-discriminative. We introduce a differentiable formulation of modularity that enables gradient-based optimization, and integrate random walk-based propagation with attention to refine unlabeled node embeddings. 

\subsection{Problem Setting}

Let $\gG$ be a simple, undirected graph with nodes $V$, edges $E$, and adjacency matrix $\mA$. Let the degree of a node $v\in V$ be $d_v$, and let the vector of degrees be $\vd$. Also let $m=|E|$ and $n=|V|$. Consider the classification task, where each node $v\in V$ is associated with a label $y_v\in\mathcal{C}$.

Let us choose an embedding dimensionality $k\in\sN$. Consider an arbitrary downstream classification model $f:\sR^k\rightarrow\mathcal{C}$. Our objective is to learn an embedding map $p:V\rightarrow\sR^k$ such that performance of the downstream task $f\circ p$ is maximized. We will learn $p$ as a continuous embedding matrix $\rmS\in\sR^{n\times k}$, where each row $\rmS_{i,:}$ denotes the $k \ll n$-dimensional embedding of node $i$, i.e., $\rmS_{i,:}=p(i)$.

\subsection{Linear Modularity Optimization}

We want to model modularity-aware embedding generation for graphs with unknown features with the matrix $\rmS$.
The modularity function can be equivalently written as
\begin{equation}
Q(\rmS) = \frac{1}{2m} \sum_{i,j} 
\left( A_{ij} - \frac{d_i d_j}{2m} \right) \mathbf{s}_i^\intercal \mathbf{s}_j,
\end{equation}
which in matrix form reduces to
\begin{equation}
Q(\rmS) = \frac{1}{2m} \mathrm{Tr}\left(\rmS^\intercal \mB \rmS\right),
\end{equation}
where $\mB = \mA - \frac{\vd \vd^\intercal}{2m}$ is the modularity matrix \citep{Modularity}.  

\paragraph{Gradient Approximation.}
Differentiating w.r.t. $\rmS$ yields
\begin{equation}
\nabla_{\rmS} Q_{\text{exact}} = \frac{1}{m} \left( \mA \rmS - \frac{1}{2m} \vd (\vd^\intercal \rmS) \right).
\end{equation}

However, for enhanced numerical stability and computational efficiency, we employ the following gradient approximation:
\begin{equation}
\nabla_{\rmS} Q_{\text{prop}} = \frac{1}{2m} \left( \mA \rmS - \frac{1}{2m} \vd (\mathbf{1}^\intercal \rmS) \right),
\label{eq:proposed_grad}
\end{equation}
where $\mathbf{1}^\intercal \rmS = \sum_i \rmS_{i,:}$ is the unweighted sum of all node embeddings.

\paragraph{Interpretation.}
The proposed gradient has an intuitive interpretation:
\begin{itemize}
    \item The term $\mA \rmS$ performs a local aggregation, where each node's embedding is updated by summing the embeddings of its neighbors. This pulls nodes towards the center of their immediate community.
    \item The term $\frac{1}{2m} \vd (\mathbf{1}^\intercal \rmS)$ acts as a global correction. It estimates the expected connection strength under the configuration model but uses the unweighted global average embedding $\frac{1}{2m}\mathbf{1}^\intercal \rmS$ instead of the degree-weighted average. This pushes nodes away from the global center of the graph, enhancing the separation between communities.
    \item The factor $\frac{1}{2m}$ scales the entire expression to be comparable across graphs of different sizes.
\end{itemize}
This approximation replaces the degree-weighted mean $\vd^\intercal \rmS$ in the exact gradient with the unweighted mean $\mathbf{1}^\intercal \rmS$. This simplifies the computation and often leads to more stable optimization, as it reduces the influence of high-degree nodes (hubs) on the global correction term, preventing their features from overly dominating the global statistics.

\paragraph{Computational Complexity.}

The main steps of sparse matrix multiplication \(\mA\rmS\) and degree corrections scale as $O(|E|k + nk)$ ($|E|$ being the number of edges), while supervised gradient updates remain linear in the number of nodes. The additional semi-supervised components add costs of  $O(w\ell)$ for $w$ random walks each of length $l$, and $O(nd_{\max}k)$ for attention updates, $d_{\max}$ being the maximum possible degree of a node. Thus, the overall complexity is $O(|E|k + nk + nd_{\max}k + w\ell)$, which is far more scalable than spectral methods requiring \(O(n^3)\) for eigen-decomposition.

\subsection{Supervised and Semi-Supervised Components}

While modularity optimization preserves structural properties, it does not enforce label consistency. We therefore introduce supervised and semi-supervised components.

\paragraph{Supervised regularization.}  
Given a set of ground-truth labels $\vy \in \sR^n$, we minimize intra-class embedding variance by defining the loss
\begin{equation}
Q_{\text{sup}} = \sum_{c} \sum_{i \in C_c} \| \rmS_{i,:} - \vmu_c \|^2,
\end{equation}
where $\vmu_c = \frac{1}{|C_c|} \sum_{i \in C_c} \rmS_{i,:}$ is the class mean. The gradient is
\begin{equation}
\nabla Q_{\text{sup}} = \rmS - \tilde{\rmS}, \quad \tilde{\rmS}_i = \bm{\mu}_c \; \text{for } i \in C_c.
\end{equation}
This ensures embeddings of labeled nodes in the same class remain clustered.  

\paragraph{Semi-supervised label propagation.}  
For unlabeled nodes, we employ biased random walks \citep{label_propagation} that preferentially visit labeled nodes, allowing labels to diffuse across the network. At each step, if labeled neighbors exist they are selected with higher probability, otherwise the walk proceeds uniformly. Repeated walks per node accumulate labeled visits, defining a propagation distribution. We will denote each labeled random walk by $\mathcal{W}$.

To refine this signal, we adopt an attention mechanism \citep{GAT,node_attention}, which weights the contribution of labeled nodes by similarity. For an unlabeled node $i$ with embedding $\rmS_{i,:}$, the attention weight for node $j$ is
\begin{equation}
w_{ij} = \frac{\exp(\rmS_{i,:}^\intercal \rmS_{j,:})}{\sum_{k \in \rho(i)} \exp(\rmS_{i,:}^\intercal \rmS_{k,:})},
\end{equation}
where $\rho(i)$ denotes the set of nodes visited in random walks from $i$. The corresponding semi-supervised gradient is
\begin{equation}
\nabla_{\rmS} Q_{\text{semi}} = \rmS_{i,:} - \sum_{j} w_{ij} \rmS_{j,:}.
\end{equation}
This encourages unlabeled embeddings to shift toward weighted averages of similar labeled neighbors.

\subsection{Optimization}

We integrate modularity, supervised, and semi-supervised objectives into a unified gradient ascent update:
\begin{equation}
\nabla_\rmS Q_{\text{total}} = \nabla_\rmS Q_{\text{prop}} - \lambda_{\text{sup}} \nabla_\rmS Q_{\text{sup}} - \lambda_{\text{semi}} \nabla_\rmS Q_{\text{semi}}.
\end{equation}
Embeddings are updated as
\begin{equation}
\rmS \leftarrow \rmS + \eta \nabla_\rmS Q_{\text{total}},
\end{equation}
where $\eta$ is the learning rate. To ensure stability, $\rmS$ is orthonormalized after each iteration via QR decomposition.  The overall procedure is represented in Algorithm~\ref{algo:overall}. Further implementation details can be found in Appendices~\ref{app:algorithm details} and \ref{app:extended results}.

\begin{algorithm}[htb!]
\caption{FUSE}
\begin{algorithmic}[1]
\Require Graph $\mathcal{G}(V,E)$, Labels $\mathbf{y}$, Label Mask $\mathbf{M}$, Learning Rate $\eta$, Regularization $\lambda_{\text{sup}}, \lambda_{\text{semi}}$, Iterations $T$
\Ensure Optimized Embeddings $\mathbf{S}$
\State Convert $\mathcal{G}$ to adjacency $\mA$, compute degrees $\vd$, total edges $m$
\State Initialize $\mathbf{S}$ randomly, orthonormalize using QR
\State $\mathcal{W} \gets \textsc{LabeledRandomWalks}(\mathcal{G}, \mathbf{M}, \mathbf{y})$\Comment{From Algorithm~\ref{algo:randomWalks}}
\State $\mathbf{W} \gets \textsc{ComputeAttentionWeights}(\mathbf{S}, \mathcal{W})$\Comment{From Algorithm~\ref{algo:attention}}
\For{$t = 1$ to $T$}
    \State $\nabla_{\mathbf{S}} Q_{\text{prop}} \gets \tfrac{1}{2m} \left( \mA \rmS - \tfrac{1}{2m} \vd (\mathbf{1}^\intercal \rmS) \right)$ \Comment{Modularity gradient}
    \State $\nabla_{\mathbf{S}} Q_{\text{sup}} \gets \mathbf{S} - \tilde{\mathbf{S}}$ \Comment{Supervised gradient} 
    \State $\nabla_{\mathbf{S}} Q_{\text{semi}} \gets \mathbf{S}_{i,:} - \sum_j w_{ij} \mathbf{S}_{j,:}$ \Comment{Semi-supervised gradient}
    \State $\mathbf{S} \gets \mathbf{S} + \eta \big( \nabla_{\mathbf{S}} Q_{\text{prop}} - \lambda_{\text{sup}} \nabla_{\mathbf{S}} Q_{\text{sup}} - \lambda_{\text{semi}} \nabla_{\mathbf{S}} Q_{\text{semi}} \big)$
    \State Orthonormalize $\mathbf{S}$ using QR-decomposition
\EndFor
\State \Return $\mathbf{S}$
\end{algorithmic}
\label{algo:overall}
\end{algorithm}
  
\section{Experiments}

\subsection{Datasets}

The evaluation of the proposed semi-supervised modularity-based node embedding method is conducted on six benchmark datasets: Cora \citep{cora}, Citeseer \citep{citeseer}, WikiCS \citep{wikics}, Amazon Photo \citep{photo}, PubMed \citep{pubmed}, and ArXiV \citep{hu2020open}. Each dataset consists of nodes representing entities and edges signifying relationships (Table~\ref{tab:dataset_stats}). For experiments, whenever necessary we mask labels of subsets of nodes (which are used for testing node classification). The experiments assume that node features are unavailable, except for the case of a trivial baseline described in Section \ref{sec:baselines}.

\subsection{Baselines}\label{sec:baselines}

We evaluate our approach against a range of baselines spanning unsupervised, self-supervised, semi-supervised and trivial embedding strategies:  
\begin{itemize}
    \item \textbf{Unsupervised baselines.} We use Node2Vec~\citep{node2vec} and DeepWalk~\citep{deepwalk}, both random-walk-based methods that employ the Skip-Gram model for representation learning. In addition, we include Variational Graph Auto-Encoders (VGAE) \newline ~\citep{vgae} as a neural network based unsupervised embedding method. For VGAE, we initialized the feature matrix as an identity matrix since we assume that features were unavailable, as recommended by \cite{vgae}. We also implemented M-NMF~\citep{wang2017m} for generating the $k$ dimensional embeddings, observing the downstream classification results later.
    \item \textbf{Self-supervised baseline.} We employ Deep Graph Infomax (DGI) \newline ~\citep{dgi}, which maximizes mutual information between node-level and graph-level representations. Here we initialized the feature matrix as a random $n\times k$ matrix ($n=$ number of nodes and $k=150$) to compare with FUSE, since we assume that features were unavailable.
    \item \textbf{Semi-supervised baseline.} We employ GraFN~\citep{lee2022grafn} and ReVAR~\citep{yan2023revar} under the \textit{non-availability of features} setting, using random feature matrices. Both frameworks combine a GNN encoder with a classifier via customized losses, making embedding generation and classification degenerate or inseparable. Hence, we tested them with different encoders (GCN, GAT, GraphSAGE). As ReVAR targets imbalanced node classification, we adapted it to the non-imbalanced case to generate embeddings through the encoders and evaluate classifier performance. Reported runtime is the sum of embedding generation and classification, as both are degenerate.
    \item \textbf{Trivial baselines.} Random embeddings serve as a lower-bound baseline, while directly using the available node features act as an upper-bound benchmark.  
\end{itemize}

Embeddings generated by each method are subsequently used as input to three GNN classifiers: GCN~\citep{GCN}, GAT~\citep{GAT}, and GraphSAGE~\citep{GraphSAGE}. For all baselines, unless otherwise mentioned we have assumed the default parameter values for all experiments. To ensure comparability, we fix the embedding dimension to $150$ and maintain identical neural architectures across datasets: a two-layer vanilla GNN or MLP with no additional hyperparameter tuning. For our method, the initialization of the embedding matrix $\rmS$ is random, and dataset-specific parameter values of FUSE are summarized in Table \ref{tab:hyperparams} in Appendix \ref{app:algorithm details}.

\subsection{Results} \label{sec:results}

We now present the empirical evaluation of our proposed method across six benchmark datasets. Results are structured around five key aspects: (1) downstream classification performance and runtime efficiency, (2) ablation studies analyzing the contributions of unsupervised, semi-supervised, components of the 
FUSE objective, (3) FUSE parameter sensitivity analysis, (4) scalability outcomes and (5) missingness experiments across different masking mechanisms.

\subsubsection{Downstream Classification Performance}

Tables \ref{tab:merged_results} summarize the classification accuracy and F1-scores obtained when embeddings from different methods are fed into GCN, GAT, and GraphSAGE under both 70-30 and 30-70 train-test splits. Several consistent trends emerge:

\begin{itemize}
\item \textbf{FUSE achieves the best or second-best performance across classifiers.} On both splits, FUSE performs at par than DeepWalk, Node2Vec and clearly outperforms self supervised algorithms like DGI along with unsupervised M-NMF and semi-supervised GraFN and ReVAR in nearly all cases. Similar to Node2Vec and DeepWalk it is robust across classifiers and also matches or even surpasses the classification performance of the given embedding. 
\item \textbf{FUSE facilitates superior learning for GCNs.} FUSE-generated embeddings especially enhance the learning capability of the GCN classifier. This is an important aspect in the context of speed and scalability since GCN is significantly faster that GAT or GraphSAGE.
\end{itemize}

Overall, these results confirm that generating embeddings via FUSE lead to strong downstream classification without requiring precomputed features.

\subsubsection{Runtime Efficiency}

Tables~\ref{tab:runtimes} and~\ref{tab:semi_supervised_performance} report embedding generation times across datasets. Although DeepWalk and Node2Vec achieve downstream classification performance at par with FUSE, our algorithm exhibits a significant computational advantage, being approximately $5$ times faster on average. This advantage is further supported by scalability studies on the ArXiV dataset (Appendix \ref{app:scalability}, Tables~\ref{tab:arxiv_embeddings_30_70} and \ref{tab:arxiv_embeddings_70_30}), where FUSE is more than $7$ times faster.

To address the potential concern that the default \texttt{walk\_length} of 80 for Node2Vec and DeepWalk might inflate their runtimes, we conducted an additional experiment with a reduced \texttt{walk\_length} of $5$ for a single seed for these two algorithms only. Interestingly across datasets, we observed that, performance remained comparable to that with the longer walk, and runtimes did improve significantly. Nonetheless, for larger datasets, especially with more edges, like Photos, WikiCS, and ArXiV (see Appendix~\ref{app:scalability}, Tables~\ref{tab:deepwalk_node2vec_results}, \ref{tab:deepwalk_node2vec_results_arxiv} and \ref{tab:runtimes_deepwalk_node2vec}), FUSE maintains its advantage, delivering superior classification performance while remaining at around $3$ times faster.

In fact, FUSE is faster than all compared unsupervised and self-supervised embedding algorithms except DGI, which performs poorly in downstream classification under the assumption of feature unavailability. Semi-supervised algorithms like GraFN and ReVAR, while computationally feasible, display significantly lower performance than Node2Vec, DeepWalk, and FUSE (Table~\ref{tab:merged_results}).

Execution times for our ablation variants are compared in Table~\ref{tab:execution_times_ours}. The semi-supervised modularity-based embeddings are only marginally slower than the purely unsupervised versions but are significantly more effective (see Table \ref{tab:unsupervised_summary}), confirming that label propagation is an efficient and beneficial addition.

\begin{table}[htb!]
\centering
\scriptsize
\begin{tabular}{ll|cc|cc}
\toprule
\multirow{2}{*}{\bf Classifier} & \multirow{2}{*}{\bf Embedding} 
& \multicolumn{2}{c|}{\bf 70-30 Split} & \multicolumn{2}{c}{\bf 30-70 Split} \\
& & \bf Accuracy & \bf F1 & \bf Accuracy & \bf F1 \\
\midrule
\multirow{8}{*}{\bf GAT} 
& Random & $0.71 \pm 0.014$ & $0.68 \pm 0.016$ & $0.48 \pm 0.028$ & $0.40 \pm 0.033$ \\
& DeepWalk & $\underline{0.82 \pm 0.008}$ & $\underline{0.80 \pm 0.009}$ & $\mathbf{0.79 \pm 0.007}$ & $\mathbf{0.77 \pm 0.009}$ \\
& Node2Vec & $\underline{0.82 \pm 0.007}$ & $\underline{0.80 \pm 0.007}$ & $\mathbf{0.79 \pm 0.007}$ & $\mathbf{0.77 \pm 0.008}$ \\
& MNMF & $0.55 \pm 0.024$ & $0.52 \pm 0.026$ & $0.34 \pm 0.024$ & $0.29 \pm 0.022$ \\
& VGAE & $0.81 \pm 0.009$ & $0.79 \pm 0.010$ & $0.78 \pm 0.005$ & $0.76 \pm 0.005$ \\
& DGI & $0.59 \pm 0.073$ & $0.51 \pm 0.098$ & $0.54 \pm 0.070$ & $0.45 \pm 0.100$ \\
& GraFN &$0.76 \pm 0.012$ & $0.71 \pm 0.052$ & $0.70 \pm 0.011$ & $0.60 \pm 0.075$ \\
& ReVAR &$0.43 \pm 0.023$ & $0.29 \pm 0.029$ & $0.42 \pm 0.017$ & $0.29 \pm 0.029$ \\
& FUSE & $\mathbf{0.82 \pm 0.009}$ & $\mathbf{0.80 \pm 0.009}$ & $\underline{0.78 \pm 0.006}$ & $\underline{0.76 \pm 0.008}$ \\
& Given Emb. & $0.86 \pm 0.005$ & $0.84 \pm 0.006$ & $0.84 \pm 0.004$ & $0.82 \pm 0.006$ \\
\midrule
\multirow{8}{*}{\bf GCN} 
& Random & $0.49 \pm 0.031$ & $0.45 \pm 0.030$ & $0.37 \pm 0.032$ & $0.33 \pm 0.028$ \\
& DeepWalk & $\underline{0.64 \pm 0.039}$ & $\underline{0.58 \pm 0.050}$ & $\underline{0.67 \pm 0.027}$ & $\underline{0.61 \pm 0.039}$ \\
& Node2Vec & $0.64 \pm 0.042$ & $0.57 \pm 0.058$ & $0.66 \pm 0.026$ & $0.61 \pm 0.036$ \\
& MNMF & $0.46 \pm 0.044$ & $0.37 \pm 0.051$ & $0.36 \pm 0.032$ & $0.29 \pm 0.026$ \\
& VGAE & $\underline{0.71 \pm 0.017}$ & $\underline{0.68 \pm 0.022}$ & $\underline{0.69 \pm 0.017}$ & $\underline{0.66 \pm 0.017}$ \\
& DGI & $0.30 \pm 0.026$ & $0.12 \pm 0.049$ & $0.32 \pm 0.048$ & $0.15 \pm 0.081$ \\
& GraFN &$0.74 \pm 0.010$ & $0.72 \pm 0.009$ & $0.66 \pm 0.006$ & $0.64 \pm 0.007$ \\
& ReVAR &$0.35 \pm 0.019$ & $0.18 \pm 0.028$ & $0.35 \pm 0.017$ & $0.18 \pm 0.028$ \\
& FUSE & $\mathbf{0.78 \pm 0.014}$ & $\mathbf{0.76 \pm 0.013}$ & $\mathbf{0.73 \pm 0.020}$ & $\mathbf{0.71 \pm 0.017}$ \\
& Given Emb. & $0.58 \pm 0.022$ & $0.49 \pm 0.018$ & $0.56 \pm 0.023$ & $0.47 \pm 0.018$ \\
\midrule
\multirow{8}{*}{\bf SAGE} 
& Random & $0.56 \pm 0.018$ & $0.51 \pm 0.015$ & $0.35 \pm 0.018$ & $0.26 \pm 0.014$ \\
& DeepWalk & $\mathbf{0.81 \pm 0.011}$ & $\mathbf{0.79 \pm 0.012}$ & $\mathbf{0.78 \pm 0.008}$ & $\mathbf{0.76 \pm 0.009}$ \\
& Node2Vec & $\mathbf{0.81 \pm 0.010}$ & $\mathbf{0.79 \pm 0.009}$ & $\underline{0.77 \pm 0.007}$ & $\underline{0.75 \pm 0.008}$ \\
& MNMF & $0.52 \pm 0.016$ & $0.47 \pm 0.021$ & $0.33 \pm 0.019$ & $0.27 \pm 0.022$ \\
& VGAE & $\underline{0.80 \pm 0.009}$ & $\underline{0.78 \pm 0.011}$ & $0.76 \pm 0.010$ & $0.74 \pm 0.011$ \\
& DGI & $0.57 \pm 0.054$ & $0.48 \pm 0.088$ & $0.54 \pm 0.047$ & $0.46 \pm 0.070$ \\
& GraFN &$0.67 \pm 0.010$ & $0.63 \pm 0.010$ & $0.55 \pm 0.008$ & $0.51 \pm 0.010$ \\
& ReVAR &$0.25 \pm 0.009$ & $0.15 \pm 0.006$ & $0.24 \pm 0.005$ & $0.16 \pm 0.006$ \\
& FUSE & $\underline{0.80 \pm 0.012}$ & $\underline{0.77 \pm 0.013}$ & $0.75 \pm 0.008$ & $0.73 \pm 0.010$ \\
& Given Emb. & $0.85 \pm 0.008$ & $0.83 \pm 0.012$ & $0.83 \pm 0.006$ & $0.80 \pm 0.008$ \\
\bottomrule
\end{tabular}
\caption{Classification accuracy and F1-score (mean $\pm$ standard deviation) across embedding methods and three classifiers for all the datasets (except ArXiV). Results are reported for both 70-30 and 30-70 train-test splits. Best and second-best (excluding given embeddings) are highlighted in \textbf{bold} and \underline{underlined}, respectively.}
\label{tab:merged_results}
\end{table}

\begin{table}[htbp!]
\centering
\scriptsize
\begin{tabular}{lcccccc}
\toprule
\textbf{Embedding} & \textbf{Cora} & \textbf{CiteSeer} & \textbf{Amazon Photo} & \textbf{WikiCS} & \textbf{PubMed} & \textbf{Average} \\
\midrule
\multicolumn{7}{c}{\textbf{70-30 Split}} \\
\midrule
Random      & 0.01 & 0.01 & 0.01 & 0.03 & 0.04 & 0.02 \\
DeepWalk    & 50.48 & 51.41 & 292.30 & 747.20 & 490.72 & 326.422 \\
Node2Vec    & 47.26 & 50.32 & 288.33 & 745.33 & 453.74 & 316.996 \\
MNMF       & 41.75 & 56.34 & 323.31 & 672.46 & 1742.94 & 567.36 \\
VGAE        & 12.95 & 14.32 & 137.28 & 329.46 & 235.24 & 145.850 \\
DGI         & \textbf{6.78} & \textbf{7.96} & \underline{53.42} & \underline{134.58} & \textbf{39.43} & \textbf{48.434} \\
FUSE       & \underline{12.52} & \underline{13.36} & \textbf{49.47} & \textbf{86.45} & \underline{95.79} & \underline{51.518} \\

\midrule
\multicolumn{7}{c}{\textbf{30-70 Split}} \\
\midrule
Random      & 0.01 & 0.01 & 0.01 & 0.03 & 0.04 & 0.02 \\
DeepWalk    & 50.99 & 51.84 & 292.98 & 792.11 & 477.77 & 333.138 \\
Node2Vec    & 47.49 & 50.65 & 290.95 & 785.70 & 448.58 & 324.674 \\
MNMF       & 41.75 & 56.34 & 323.31 & 672.46 & 1742.94 & 567.36 \\
VGAE        & 12.97 & 14.48 & 136.07 & 338.10 & 226.29 & 145.582 \\
DGI         & \textbf{6.83} & \textbf{7.33} & \textbf{53.37} & \textbf{126.80} & \textbf{36.05} & \textbf{46.076} \\
FUSE       & \underline{14.42} & \underline{14.31} & \underline{64.55} & \underline{128.92} & \underline{109.15} & \underline{66.27} \\

\bottomrule
\end{tabular}
\caption{Runtime comparison (in seconds) of different embedding methods across datasets (except ArXiV) under 70-30 and 30-70 train-test splits. Reported values are averages over $5$ runs. Best and second-best (excluding random embeddings) are highlighted in \textbf{bold} and \underline{underlined}, respectively.}
\label{tab:runtimes}
\end{table}

\subsubsection{Additional Analyses}

To substantiate the effectiveness and robustness of FUSE, we conducted ablation, sensitivity, scalability and masking studies (details in Appendix~\ref{app:extended results}).  

\textbf{Ablation Study.} We evaluated the individual contributions of the semi-supervised and unsupervised objectives (Appendix~\ref{app:ablation study}), as well as their combination, under both the 30-70 and 70-30 train-test splits (assumed learning rate $0.05$). The unsupervised component of the FUSE objective alone performs significantly well compared to the only semi-supervised counterpart especially for the GraphSAGE classifier (Tables~\ref{tab:unsupervised_summary} and \ref{tab:unsupervised_summary_70_30}). This indicates that FUSE can also adapt well to scenarios where labels are completely unavailable using only the modularity-driven objective. The semi-supervised component alone is also at-par with the unsupervised component in terms of classification performance. However, the unsupervised objective alone proves to be faster (Tables~\ref{tab:execution_times_ours} and \ref{tab:execution_times_ours_70_30}). It is clear from the overall results however, that incorporating all three components of the objective is indeed advantageous, especially for large scale datasets.

\textbf{Sensitivity Analysis.} We analyzed robustness to hyperparameters (Tables~\ref{tab:all_optimal_hyperparams} (a, b, c)). Learning rate $\eta$ and loss weights $\lambda_{\text{sup}}, \lambda_{\text{semi}}$ were most sensitive, while structural parameters ($r$, $L,L'$) tolerated wider ranges. Deeper settings sometimes improved accuracy but increased runtime disproportionately, suggesting moderate configurations as optimal (Appendix~\ref{app:sensitivity}).

\textbf{Scalability Experiments.}  We additionally evaluated FUSE on a large-scale graph \arxivlink ~ to assess its applicability to real-world settings. The results and execution times are reported in Tables~\ref{tab:arxiv_embeddings_30_70} and \ref{tab:arxiv_embeddings_70_30} (Appendix~\ref{app:scalability}). 

\textbf{Label masking Experiments.} In real-world datasets, class distributions among unlabeled nodes are often highly imbalanced. To assess the robustness of FUSE under such imbalance, we evaluated its performance with three label-masking strategies at 20\%, 50\%, and 80\% missingness on the Cora and CiteSeer datasets (details in Appendix~\ref{app:missingness}). FUSE remained consistently competitive across all settings, showing a particular advantage with the GCN and GAT classifiers under high missingness rates (80\%) and more challenging masking schemes (MAR, MNAR).

\section{Discussion and conclusion} \label{sec:discussion_and_conclusion}

In this paper we introduce FUSE, a fast, scalable and high-performance node embedding generation algorithm that does not require predefined features. The objective function of FUSE integrates an unsupervised, a semi-supervised and a supervised component. 

The unsupervised component of FUSE objective is based on a novel linear-time maximization of graph modularity, which enables runtime and performance-efficient embedding generation even in the absence of labels. Modularity being a global graph property, this component can be interpreted as learning global structural features. The semi-supervised component, on the other hand, leverages label-biased random walks and inter-node attention between labeled and unlabeled nodes. This allows the model to capture local structures at the node or neighborhood level during feature learning. Supported by the global structure learning of the unsupervised module, we observe that FUSE can extract meaningful local features using short random walks of length as little as five. This also contributes to the overall runtime efficiency of FUSE. Finally, the supervised component reduces intra-class embedding variance, ensuring that nodes belonging to the same class are closely aligned in the embedding space. By combining these elements, FUSE achieves accuracy comparable to or better than established baselines, while being five to seven times faster, particularly on large-scale datasets such as ArXiv.

Nevertheless, FUSE has some limitations that we would like to highlight. FUSE is designed to operate in settings where node features are assumed to be unavailable. It is thus unable to incorporate information extraneous to the graph structure. A simple extension of the algorithm 
to incorporate node features would be to concatenate these features onto the embedding matrix $\rmS$. 
Another interesting avenue would be to explore this framework in dynamically updating graphs while maintaining scalability.

In practice, FUSE being considerably faster and performance-efficient, it can be applicable to resource-constrained or dynamic environments. In dynamic environments, for example, the embeddings can be updated more frequently. The ability to operate effectively with sparse or missing labels further enhances its practicality in real-world domains.



\section*{Author Contributions}

Sujan Chakraborty developed the core algorithm and conducted the experimental analyses. Rahul Bordoloi contributed to the benchmarking studies and assisted in refining the methodology. Saptarshi Bej conceptualized the research framework and guided the methodological design. Olaf Wolkenhauer and Anindya Sengupta supervised the overall study and provided critical feedback and insights throughout the research process. The authors declare that they have no conflict of interest.

\bibliography{iclr2026_conference}
\bibliographystyle{plainnat}

\appendix

\section{Dataset and Algorithm Details and Visualizations}
\label{app:algorithm details}

\begin{figure}[htbp!]

    \noindent
\vspace*{-0.2cm} 
\begin{tikzpicture}[
  scale=0.7, transform shape, 
  >=Stealth,
  every node/.style={font=\footnotesize}, 
  box/.style={
    draw, rounded corners=2mm, thick, align=center,
    minimum height=6mm, inner sep=1.5mm, text width=0.18\linewidth 
  },
  input/.style={box, fill=green!20},
  setup/.style={box, fill=blue!15},
  loopheader/.style={box, fill=orange!20},
  loopbody/.style={box, fill=yellow!15},
  output/.style={box, fill=green!20},
  arrow/.style={-Stealth, thick}
]

\matrix (M) [matrix of nodes, nodes={box}, column sep=4mm, row sep=3mm]{
  & |[input, text width=0.4\linewidth]| {Input: Graph $\gG=(V,E)$, labels $y_L$, dim $k$, params $(\eta,\lambda_{\mathrm{sup}},\lambda_{\mathrm{semi}},T)$} & \\
  |[setup, text width=0.25\linewidth]| {Initialize $\rmS \in \sR^{|V|\times k}$; \\ Orthonormalize (QR)} & & |[setup, text width=0.25\linewidth]| {Label-aware random walks / neighborhoods} \\
  & |[setup, text width=0.3\linewidth]| {Compute attention / weights} & \\
  & |[loopheader]| {Loop: $t=1,\dots,T$} & \\
  |[loopbody]| {$\nabla_{\mathrm{mod}}$ \\ (modularity)} & |[loopbody]| {$\nabla_{\mathrm{sup}}$ \\ (supervised)} & |[loopbody]| {$\nabla_{\mathrm{semi}}$ \\ (semi-supervised)} \\
  & |[loopbody, text width=0.38\linewidth]| {Update $\rmS$: $\rmS \leftarrow \rmS + \eta\!\left(\nabla_{\mathrm{mod}} - \lambda_{\mathrm{sup}}\nabla_{\mathrm{sup}} - \lambda_{\mathrm{semi}}\nabla_{\mathrm{semi}}\right)$ \\ Re-orthonormalize (QR)} & \\
  & |[output, text width=0.3\linewidth]| {Output: final embeddings $\rmS$} & \\
};

\draw[arrow] (M-1-2) -| (M-2-1);
\draw[arrow] (M-1-2) -| (M-2-3);
\draw[arrow] (M-2-3) |- (M-3-2);
\draw[arrow] (M-2-1) |- (M-4-2);
\draw[arrow] (M-3-2) -- (M-4-2);

\draw[arrow] (M-4-2) -- (M-5-1);
\draw[arrow] (M-4-2) -- (M-5-2);
\draw[arrow] (M-4-2) -- (M-5-3);

\draw[arrow] (M-5-1) -- (M-6-2);
\draw[arrow] (M-5-2) -- (M-6-2);
\draw[arrow] (M-5-3) -- (M-6-2);

\draw[arrow] (M-6-2) -- (M-7-2);
\end{tikzpicture}
    \caption{Algorithm pipeline (FUSE).}
    \label{fig:flowcharts:algo}

    \centering
    \begin{center}
\resizebox{0.95\linewidth}{!}{
\begin{tikzpicture}[
  >=Stealth,
  every node/.style={font=\footnotesize}, 
  box/.style={
    draw, rounded corners=2mm, thick, align=center,
    minimum height=7mm, inner sep=2mm, text width=0.25\linewidth
  },
  dataset/.style={box, fill=green!15},
  preprocess/.style={box, fill=blue!15},
  split/.style={box, fill=orange!15},
  embed/.style={box, fill=purple!15},
  train/.style={box, fill=red!15},
  evaluate/.style={box, fill=brown!15},
  results/.style={box, fill=green!25, text width=0.24\linewidth, minimum height=9mm},
  arrow/.style={-Stealth, thick, shorten >=1pt, shorten <=1pt}
]

\node[dataset]   (datasets)   at (0,2.5)   {Datasets \\ (Cora, Citeseer, PubMed, WikiCS,\\ Amazon Photo, ArXiV)};
\node[preprocess](preprocess) at (4,2.5) {Preprocess \\ (build adjacency $\mA$)};
\node[split]     (split)      at (8,2.5)   {Create splits \\ (train / test)};

\node[embed]     (embed)      at (0,0)     {Compute embeddings $\mathrm{\rmS}$ \\ with embedding algorithms};
\node[train]     (train)      at (4,0)   {Train on $\mathrm{\rmS}$ \\ (GCN/GAT/SAGE)};
\node[evaluate]  (evaluate)   at (8,0)     {Evaluate \& analyze \\ Accuracy, F1, runtime; ablation};

\node[results]   (results)    at (4,-1.5) {Results \& comparisons};

\draw[arrow] (datasets.east) -- (preprocess.west);
\draw[arrow] (preprocess.east) -- (split.west);

\draw[arrow] (split.south) -- ++(0,-0.8) -| (embed.north);

\draw[arrow] (embed.east) -- (train.west);
\draw[arrow] (train.east) -- (evaluate.west);

\draw[arrow] (evaluate.south) |- (results.east);

\end{tikzpicture}
} 
\end{center}
    \caption{Experimental workflow.}
    \label{fig:flowcharts:exp}

  \label{fig:flowcharts}
\end{figure}

\begin{algorithm}[htbp!]
\caption{Labeled Random Walks}
\begin{algorithmic}[1]
\Require Graph $\mathcal{G}(V,E)$, Label Mask $\mathbf{M}$, Labels $\mathbf{y}$, Walks per node $r$, Walk length $L$, Labeled step limit $L'$
\Ensure Labeled Walk Set $\mathcal{W}$
\For{each node $i \in V$}
    \For{$j = 1$ to $r$}
        \State Start walk at node $i$, set labeled count $=0$
        \For{$t = 1$ to $L$}
            \State Pick next node $j$ uniformly from neighbors of current node
            \If{$j$ is labeled \textbf{and} labeled count $< L'$}
                \State Increment labeled count
            \EndIf
            \State Store $j$ if labeled
        \EndFor
    \EndFor
\EndFor
\State \Return $\mathcal{W}$
\end{algorithmic}
\label{algo:randomWalks}
\end{algorithm}

\begin{algorithm}[htbp!]
\caption{Compute Attention Weights}
\begin{algorithmic}[1]
\Require Embeddings $\rmS$, Labeled Walks $\mathcal{W}$
\Ensure Attention Weights $W$
\For{each unlabeled node $i \in V$}
    \For{each labeled node $j \in \mathcal{W}[i]$}
        \State Compute similarity: $s_{ij} = \rmS_{i,:}^\intercal \rmS_{j,:}$
        \State Compute attention: $w_{ij} = \dfrac{\exp(s_{ij})}{\sum_{k \in \mathcal{W}[i]} \exp(s_{ik})}$
    \EndFor
\EndFor
\State \Return $W$
\end{algorithmic}
\label{algo:attention}
\end{algorithm}

\begin{table}[htbp!]
\centering
\scriptsize
\begin{tabular}{lcl}
\toprule
\textbf{Parameter} & \textbf{Value} & \textbf{Description} \\
\midrule
$k$ & $150$ & Learnt node embedding dimension (in case node embeddings are not given)\\
$\eta$ & $0.05$ & Learning rate \\
$\lambda_{\text{supervised}}$ & $1.0$ & Supervised loss weight \\
$\lambda_{\text{semi-supervised}}$ & $2.0$ & Semi-supervised loss weight \\
$T$ & $200$ & Number of gradient ascent iterations \\
$r$ & $10$ & Number of random walks per node \\
$L$ & $5$ & Length of each random walk \\
$L'$ & $3$ & Maximum labeled steps in a walk \\
\bottomrule
\end{tabular}
\caption{Hyperparameters used in semi-supervised modularity optimization for all datasets.}
\label{tab:hyperparams}
\end{table}

\begin{table}[htbp!]
    \centering
    \small
    \begin{tabular}{lcccc}
        \hline
         \textbf{Dataset} & \textbf{\# Nodes} & \textbf{\# Edges} & \textbf{\# Classes} & \textbf{Given Embedding Dim.} \\
        \hline
        Cora & 2,708 & 5,429 & 7 & 1,433 \\
        CiteSeer & 3,327 & 9,104 & 6 & 3,703 \\
        PubMed & 19,717 & 44,338 & 3 & 500 \\
        Amazon Photo & 7,487 & 119,043 & 8 & 745 \\
        WikiCS & 11,701 & 216,123 & 10 & 300 \\
        ArXiV & 1,69,343 & 1,166,243 & 40 & 128 \\
        \hline
    \end{tabular}
    \caption{Statistics of the benchmark datasets used in the experiments.}
    \label{tab:dataset_stats}
\end{table}

\section{Extended Results} \label{app:extended results}

\subsection{Semi-Supervised Baselines}

\begin{table}[htbp]
\centering
\begin{tabular}{ll|cc|c}
\toprule
\textbf{Model} & \textbf{Encoder (Split)} & \textbf{Accuracy} & \textbf{F1} & \textbf{Time (s)} \\
\midrule
\multirow{6}{*}{GraFN}
    & GCN (70-30)   & $0.74 \pm 0.010$ & $0.72 \pm 0.009$ & 18.65 \\
    & GCN (30-70)   & $0.66 \pm 0.006$ & $0.64 \pm 0.007$ & 18.64 \\
    & GAT (70-30)   & $0.76 \pm 0.012$ & $0.71 \pm 0.052$ & 103.80 \\
    & GAT (30-70)   & $0.70 \pm 0.011$ & $0.60 \pm 0.075$ & 103.83 \\
    & SAGE (70-30)  & $0.67 \pm 0.010$ & $0.63 \pm 0.010$ & 10.89 \\
    & SAGE (30-70)  & $0.55 \pm 0.008$ & $0.51 \pm 0.010$ & 10.89 \\
\midrule
\multirow{6}{*}{ReVAR}
    & GCN (70-30)   & $0.35 \pm 0.019$ & $0.18 \pm 0.028$ & 43.74 \\
    & GCN (30-70)   & $0.35 \pm 0.017$ & $0.18 \pm 0.028$ & 43.53 \\
    & GAT (70-30)   & $0.43 \pm 0.023$ & $0.29 \pm 0.029$ & 385.26 \\
    & GAT (30-70)   & $0.42 \pm 0.017$ & $0.29 \pm 0.029$ & 378.15 \\
    & SAGE (70-30)  & $0.25 \pm 0.009$ & $0.15 \pm 0.006$ & 27.04 \\
    & SAGE (30-70)  & $0.24 \pm 0.005$ & $0.16 \pm 0.006$ & 28.67 \\
\bottomrule
\end{tabular}
\caption{Performance metrics (Accuracy, F1-score, and Execution Time in seconds) of the semi supervised baselines for all datasets (except ArXiV) across 70-30 and 30-70 splits. Values are averages over five runs.}
\label{tab:semi_supervised_performance}
\end{table}

Table~\ref{tab:semi_supervised_performance} represents the results along with the time required for each of ReVAR and GraFN. In these models, the embedding generation and classification process is degenerate for which the times reported are a combination of the two instead of just the embedders as reported in Table~\ref{tab:runtimes}.

\subsection{Ablation study}  \label{app:ablation study}

\begin{table}[htbp]
\centering
\begin{tabular}{ll|cc}
\hline
\multirow{2}{*}{\bf Classifier} & \multirow{2}{*}{\bf Loss} 
& \multirow{2}{*}{\bf Accuracy} & \multirow{2}{*}{\bf F1}\\
& & &  \\
\hline
\multirow{3}{*}{\bf GAT} 
& Only Semi-supervised Component & 0.690 & 0.657 \\
& Both components & 0.697 & 0.666 \\
& Only Unsupervised Component & 0.688 & 0.657 \\
\hline
\multirow{3}{*}{\bf GCN} 
& Only Semi-supervised Component & 0.656 & 0.633 \\
& Both components & 0.668 & 0.649 \\
& Only Unsupervised Component & 0.660 & 0.637 \\
\hline
\multirow{3}{*}{\bf SAGE} 
& Only Semi-supervised Component & 0.525 & 0.530 \\
& Both components & 0.732 & 0.707 \\
& Only Unsupervised Component & 0.716 & 0.696 \\
\hline

\end{tabular}
\caption{Classification accuracy and F1-score across different FUSE variants and classifiers on the 30-70 split averaged across datasets.}
\label{tab:unsupervised_summary}
\end{table}

\begin{table}[htbp]
\centering
\scriptsize
\begin{tabular}{lccccc|c}
\hline
\multirow{2}{*}{\bf Embedding} 
& \multicolumn{5}{c|}{\bf Dataset} & \multirow{2}{*}{\bf Average} \\
& Cora & CiteSeer & Amazon Photo & WikiCS & PubMed & \\
\hline
Only Semisupervised   &  8.18 &  8.04 &  50.18 &  79.60 &  103.62 & 49.124 \\
Only Unsupervised    &  3.44 &  4.84 &  11.04 &  24.02 &  42.63 &  17.194 \\
Both & 8.99 &  8.63 &  46.86 &  77.39 &  126.36 & 53.246 \\
\hline
\end{tabular}
\caption{Execution times (in seconds) of different FUSE components across datasets for 30-70 split.}
\label{tab:execution_times_ours}
\end{table}

\begin{table}[htbp]
\centering
\begin{tabular}{ll|cc}
\hline
\multirow{2}{*}{\bf Classifier} & \multirow{2}{*}{\bf Loss} 
& \multirow{2}{*}{\bf Accuracy} & \multirow{2}{*}{\bf F1} \\
& &  &  \\
\hline
\multirow{3}{*}{\bf GAT} 
& Only Semi-supervised Component & 0.75 & 0.72 \\
& Both components & 0.74 & 0.72 \\
& Only Unsupervised Component & 0.74 & 0.72 \\
\hline
\multirow{3}{*}{\bf GCN} 
& Only Semi-supervised Component & 0.71 & 0.68 \\
& Both components & 0.70 & 0.68 \\
& Only Unsupervised Component & 0.71 & 0.68 \\
\hline
\multirow{3}{*}{\bf SAGE} 
& Only Semi-supervised Component & 0.69 & 0.66 \\
& Both components & 0.76 & 0.73 \\
& Only Unsupervised Component & 0.75 & 0.73 \\
\hline
\end{tabular}
\caption{Classification accuracy and F1-score across different FUSE variants and classifiers on the 70-30 split averaged across datasets.}
\label{tab:unsupervised_summary_70_30}
\end{table}

\begin{table}[htbp]
\centering
\scriptsize
\begin{tabular}{lccccc|c}
\hline
\multirow{2}{*}{\bf Embedding} 
& \multicolumn{5}{c|}{\bf Dataset} & \multirow{2}{*}{\bf Average} \\
& Cora & CiteSeer & Amazon Photo & WikiCS & PubMed & \\
\hline
Only Semisupervised   &  6.74 &  7.79 &  31.04 &  53.81 &  69.69 &  33.814 \\
Only Unsupervised    &  3.49 &  4.15 &  10.73 &  23.84 &  34.66 &  15.574 \\
Both & 8.08 &  7.29 &  33.76 &  58.75 &  70.07 &  35.59 \\
\hline
\end{tabular}
\caption{Execution times (in seconds) of different FUSE components across datasets for 70-30 split.}
\label{tab:execution_times_ours_70_30}
\end{table}

To complement the analysis in the main text, we provide a more detailed view of the ablation experiments that disentangle the contributions of the semi-supervised and unsupervised components within FUSE. The learning rate of the FUSE algorithm was adjusted to $10^3$ for the \textit{Only Unsupervised Component} case. All other relevant parameter values remains the same. Tables~\ref{tab:unsupervised_summary} and \ref{tab:execution_times_ours} summarize performance and runtime, respectively, across different classifiers and datasets for the 30-70 split, while Tables~\ref{tab:unsupervised_summary_70_30} and \ref{tab:execution_times_ours_70_30} shows the same for the 70-30 split.

\textbf{Performance Across Classifiers.} As shown in Tables~\ref{tab:unsupervised_summary} and \ref{tab:unsupervised_summary_70_30}, the relative contribution of each component is consistent across GAT, GCN, and GraphSAGE. Notably, embeddings trained with only the unsupervised modularity term are better than those using only the semi-supervised term on an average. This confirms that community structure provides a strong inductive bias even when label information is sparse. However, combining both objectives consistently yields the highest accuracy and F1-scores overall, demonstrating that structural and label-based signals are complementary rather than interchangeable. Interestingly, the performance gap between ``Both'' and ``Unsupervised only'' is smaller than that between ``Both'' and ``Semi-supervised only,'' especially for GraphSAGE, suggesting that topology carries more transferable information than a small label set in these benchmarks.

\textbf{Runtime Considerations.} Tables~\ref{tab:execution_times_ours} and \ref{tab:execution_times_ours_70_30} highlights that the efficiency of FUSE is not compromised by integrating multiple objectives. The combined loss incurs only a marginal overhead relative to either component in isolation, while producing markedly better embeddings. This efficiency gain stems from the linearized modularity update, which dominates the runtime irrespective of whether label propagation is included. We also observe that datasets with larger number of nodes and denser connectivity like PubMed and WikiCS yield proportionally higher execution times, but the scaling behavior remains consistent across variants.

\begin{table*}[bp]
    \centering
    \scriptsize
    \begin{minipage}{\textwidth}
        \centering
        \begin{tabular}{lcccccccccc}
            \hline
            \textbf{Dataset} & $k$ & $\eta$ & $\lambda_{\text{sup}}$ & $\lambda_{\text{semi}}$ & $T$ & $r$ & $L$ & $L'$ & \textbf{Accuracy (\%)} & \textbf{Time (s)} \\
            \hline
            Cora         & 145 & 0.31 & 0.6 & 1.9 & 200 & 20 & 4 & 1 & 80.47 & 18.92 \\
            CiteSeer     & 135 & 0.51 & 0.8 & 1.5 & 450 & 13 & 5 & 3 & 63.32 & 25.56 \\
            PubMed       & 155 & 0.11 & 0.9 & 2.0 & 450 & 12 & 9 & 1 & 81.17 & 226.58 \\
            WikiCS       & 130 & 0.28 & 1.1 & 1.1 & 200 & 20 & 3 & 2 & 74.05 & 76.66 \\
            Amazon Photo & 100 & 0.21 & 1.7 & 2.5 & 300 & 13 & 3 & 2 & 89.15 & 59.47 \\
            \hline
        \end{tabular}
        \caption*{(a) Optimal hyperparameters in the $30$-$70$ setup.}
    \end{minipage}

    \vspace{1em}

    \begin{minipage}{\textwidth}
        \centering
        \begin{tabular}{lcccccccccc}
            \hline
            \textbf{Dataset} & $k$ & $\eta$ & $\lambda_{\text{sup}}$ & $\lambda_{\text{semi}}$ & $T$ & $r$ & $L$ & $L'$ & \textbf{Accuracy (\%)} & \textbf{Time (s)} \\
            \hline
            Cora         & 170 & 0.35 & 0.5 & 2.5 & 250 & 20 & 3 & 2 & 85.59 & 21.14 \\
            CiteSeer     & 200 & 0.79 & 2.2 & 1.1 & 300 & 15 & 4 & 3 & 73.74 & 24.75 \\
            PubMed       & 180 & 0.59 & 2.2 & 1.2 & 350 & 18 & 3 & 3 & 84.09 & 303.46 \\
            WikiCS       & 140 & 0.37 & 1.8 & 2.3 & 250 & 20 & 3 & 1 & 76.75 & 78.22 \\
            Amazon Photo & 120 & 0.79 & 2.1 & 1.1 & 100 & 12 & 4 & 3 & 90.71 & 34.42 \\
            \hline
        \end{tabular}
        \caption*{(b) Optimal hyperparameters in the $70$-$30$ setup.}
    \end{minipage}

    \vspace{1em}

    \begin{minipage}{\textwidth}
        \centering
        \begin{tabular}{lcccccccc}
            \hline
            \textbf{Dataset} & $\eta$ & $\lambda_{\text{sup}}$ & $\lambda_{\text{semi}}$ & $r$ & $L$ & $L'$ & \textbf{Accuracy (\%)} & \textbf{Time (s)} \\
            \hline
           Cora         & 0.25 & 0.9 & 2.3 & 17 & 4 & 9 & 80.16 & 21.12 \\
            CiteSeer     & 0.35 & 1.3 & 1.7 & 15 & 7 & 2 & 63.27 & 35.57 \\
            PubMed       & 0.46 & 0.8 & 2.5 & 18 & 4 & 1 & 80.93 & 79.95 \\
            WikiCS       & 0.03 & 0.8 & 1.3 & 15 & 3 & 2 & 73.79 & 71.44 \\
            Amazon Photo & 0.49 & 1.1 & 2.3 & 9 & 3 & 10 & 89.09 & 44.61 \\
            \hline
        \end{tabular}
        \caption*{(c) Optimal hyperparameters under $k{=}150$, $T{=}200$ for the 30-70 setup.}
    \end{minipage}

    \caption{Optimal hyperparameters of FUSE.}
    \label{tab:all_optimal_hyperparams}
\end{table*}

These results provide additional evidence that FUSE's strength lies not in any single component, but in their unification. The unsupervised modularity term ensures that embeddings respect community structure, while the semi-supervised propagation aligns them with available labels. Their joint optimization balances exploration of global topology with exploitation of label information, leading to robust performance without significant runtime penalties. 

\subsection{Sensitivity analysis}  \label{app:sensitivity}

To further assess the robustness of FUSE, we carried out a sensitivity analysis of its main hyperparameters across datasets. Tables~\ref{tab:all_optimal_hyperparams} (a, b, c) summarize the optimal settings discovered under two search protocols. These results provide insights into which hyperparameters consistently influence performance and which are less critical. 

\textbf{Influential Hyperparameters.} Among the parameters, the learning rate $\eta$ and the loss weights $\lambda_{\text{sup}}$ and $\lambda_{\text{semi}}$ emerge as the most sensitive across datasets. Small variations in $\eta$ often lead to pronounced differences in both accuracy and convergence speed, indicating the need for dataset-specific tuning. Similarly, the balance between the supervised and semi-supervised terms must be carefully adjusted, as an overemphasis on one can suppress the benefits of the other. By contrast, the neighborhood radius $r$ and structural depths $L,L'$ showed more stable behavior, with broad ranges yielding near-optimal accuracy.

\textbf{Consistency Across Datasets.} Interestingly, although the exact optimal values vary, the relative importance of hyperparameters remains consistent. For example, on both Cora and PubMed, adjusting $\lambda_{\text{semi}}$ within $[2,2.5]$ was essential to achieve competitive performance, while on WikiCS and CiteSeer, a more balanced weighting was required. The Amazon Photo dataset was less sensitive overall, achieving high accuracy under multiple configurations, suggesting that denser graphs with richer labels are inherently more robust to hyperparameter shifts.

\textbf{Runtime Trade-offs.} The sensitivity analysis also reveals a runtime– \newline performance trade-off. While larger values of $T$ or deeper $L,L'$ occasionally yield marginal accuracy gains, they incur disproportionately higher costs in training time (e.g., PubMed in Table~\ref{tab:all_optimal_hyperparams} (a and b). This indicates diminishing returns from overparameterization, and reinforces the practical value of moderate configurations that balance accuracy and efficiency.

\subsection{Scalability experiments} \label{app:scalability}
\begin{enumerate}
    \item \textbf{Purpose and Setup for ArXiV.} In addition to the experiments above, we performed another benchmarking experiment on a large dataset, namely \arxivlink for investigating the scalability of FUSE. This dataset is higly imbalanced as well. Given that the dataset is significantly larger than others for FUSE we have considered a learning rate of $0.05$ to ensure convergence within 200 iterations. For VGAE, we took the initial matrix to be a $n \times k$ random matrix instead of the $n \times n$ identity matrix as assuemd in other experiments. This is to avoid scalability issues due to very large value of $n$ for this dataset. All other parameters remain the same.

        \textbf{Observations on ArXiV.} The results across the two splits (30-70 and 70-30) for a fixed seed is given in Tables~\ref{tab:arxiv_embeddings_30_70} and \ref{tab:arxiv_embeddings_70_30}
        The results reveal that FUSE is not only scalable and robust to labels, but performs at par with unsupervised algorithms like Node2Vec and DeepWalk in terms of performance metrics. Furthermore it has a significant advantage in terms of computational time. In addition, it outperforms the semi supervised algorithms like GraFN and ReVAR In terms of Accuracy, F1 Score by a large margin. Notably, the MNMF algorithm was not scalable to this particular dataset.

    \item \textbf{Anaysis for Node2Vec and DeepWalk for a lower \texttt{walk\_length}.} To address the potential concern that the default \texttt{walk\_length} of 80 for Node2Vec and DeepWalk might inflate their runtimes, we conducted an additional experiment with a reduced \texttt{walk\_length} of $5$ for a single seed for these two algorithms. Tables~\ref{tab:deepwalk_node2vec_results}, \ref{tab:deepwalk_node2vec_results_arxiv}, and \ref{tab:runtimes_deepwalk_node2vec} summarize the results of this experiment across all datasets, reporting classification accuracy, F1-score, and runtime for both 70-30 and 30-70 train-test splits.  

    \textbf{Performance Analysis:} For most datasets, the classification performance of Node2Vec and DeepWalk with the shorter walk length remained largely comparable to that obtained with the default longer walk, suggesting that reducing the walk length does not severely compromise the quality of learned embeddings.
    
    \textbf{Runtime Comparison:} Reducing the \texttt{walk\_length} substantially improved the runtime of both Node2Vec and DeepWalk across datasets. As reported in Table~\ref{tab:runtimes_deepwalk_node2vec}, runtime reductions of FUSE with respect to these two algorithms are particularly significant for large datasets like Photos, WikiCS, and ArXiV with higher number of edges. For example, on ArXiV, DeepWalk and Node2Vec required approximately $3{,}100$--$3{,}200$ seconds for the 70-30 split, whereas FUSE completed within $1{,}360$ seconds which is roughly a $3$ times improvement in speed.  
    
    \textbf{FUSE Advantage:} Despite the reduction in random walk length for Node2Vec and DeepWalk, FUSE was consistently equally or more effective in both performance and runtime metrics. FUSE embeddings yielded higher classification accuracy and F1-scores compared to DeepWalk and Node2Vec especially for a larger dataset like ArXiV with higher number of edges, even when the latter used a very short walk length. This indicates that FUSE’s embedding methodology is not only scalable but also robust to variations in graph size and connectivity, offering a more efficient alternative for large-scale graph representation learning (Tables~\ref{tab:deepwalk_node2vec_results}, \ref{tab:deepwalk_node2vec_results_arxiv}).  
\end{enumerate}

\begin{table}[htbp]
\centering
\footnotesize
\begin{tabular}{ll|cc|cc}
\toprule
\multirow{2}{*}{\bf Classifier} & \multirow{2}{*}{\bf Embedding} 
& \multicolumn{2}{c|}{\bf 70-30 Split} & \multicolumn{2}{c}{\bf 30-70 Split} \\
& & \bf Accuracy & \bf F1 & \bf Accuracy & \bf F1 \\
\midrule
\multirow{2}{*}{\bf GAT} 
& DeepWalk (walk\_length=5) & 0.81 & 0.793 & \textbf{0.78} & \textbf{0.764} \\
& Node2Vec (walk\_length=5)& 0.81 & 0.792 & \textbf{0.78} & 0.756 \\
& FUSE & \textbf{0.82} & \textbf{0.795} & \textbf{0.78} & 0.751 \\
\midrule
\multirow{2}{*}{\bf GCN} 
& DeepWalk (walk\_length=5)& 0.62 & 0.552 & 0.65 & 0.578 \\
& Node2Vec (walk\_length=5)& 0.62 & 0.554 & 0.65 & 0.598 \\
& FUSE & \textbf{0.77} & \textbf{0.753} & \textbf{0.73} & \textbf{0.699} \\
\midrule
\multirow{2}{*}{\bf SAGE} 
& DeepWalk (walk\_length=5)& \textbf{0.81} & \textbf{0.786} & \textbf{0.78} & 0.\textbf{754} \\
& Node2Vec (walk\_length=5)& \textbf{0.81} & 0.781 & 0.77 & 0.751 \\
& FUSE & 0.79 & 0.769 & 0.75 & 0.731 \\
\bottomrule
\end{tabular}
\caption{Classification accuracy and F1-score (averaged) for DeepWalk (walk\_length=5), Node2Vec (walk\_length=5) and FUSE across three classifiers for all the datasets (except ArXiV) for a fixed seed. Results are reported for both 70-30 and 30-70 train-test splits.}
\label{tab:deepwalk_node2vec_results}
\end{table}

\begin{table}[htbp]
\centering
\footnotesize
\begin{tabular}{ll|cc|cc}
\toprule
\multirow{2}{*}{\bf Classifier} & \multirow{2}{*}{\bf Embedding} 
& \multicolumn{2}{c|}{\bf 70-30 Split} & \multicolumn{2}{c}{\bf 30-70 Split} \\
& & \bf Accuracy & \bf F1 & \bf Accuracy & \bf F1 \\
\midrule
\multirow{2}{*}{\bf GAT} 
& DeepWalk (walk\_length=5)& 0.66 & 0.42 & \textbf{0.65} & 0.40 \\
& Node2Vec (walk\_length=5)& 0.64 & 0.39 & 0.64 & 0.38 \\
& FUSE & \textbf{0.67} & \textbf{0.47} & 0.64 & \textbf{0.43} \\
\midrule
\multirow{2}{*}{\bf GCN} 
& DeepWalk (walk\_length=5)& 0.47 & 0.18 & 0.48 & 0.21 \\
& Node2Vec (walk\_length=5)& 0.46 & 0.15 & \textbf{0.49} & \textbf{0.22} \\
& FUSE & \textbf{0.50} & \textbf{0.24} & 0.45 & 0.14 \\
\midrule
\multirow{2}{*}{\bf SAGE} 
& DeepWalk (walk\_length=5)& 0.61 & \textbf{0.23} & \textbf{0.60} & 0.23 \\
& Node2Vec (walk\_length=5)& 0.59 & 0.22 & 0.58 & 0.21 \\
& FUSE & \textbf{0.62} & \textbf{0.23} & \textbf{0.60} & \textbf{0.25} \\
\bottomrule
\end{tabular}
\caption{Classification accuracy and F1-score for DeepWalk (walk\_length=5), Node2Vec (walk\_length=5) and FUSE across three classifiers for ArXiV for a fixed seed. Results are reported for both 70-30 and 30-70 train-test splits. The best metric values across each classifier have been highlited in \textbf{bold}.}
\label{tab:deepwalk_node2vec_results_arxiv}
\end{table}

\begin{table}[htbp]
\centering
\scriptsize
\begin{tabular}{lcccccc}
\toprule
\textbf{Embedding} & \textbf{Cora} & \textbf{CiteSeer} & \textbf{Amazon Photo} & \textbf{WikiCS} & \textbf{PubMed} & \textbf{ArXiV} \\
\midrule
\multicolumn{7}{c}{\textbf{70-30 Split}} \\
\midrule
DeepWalk (walk\_length=5)    & 3.92 & 4.23 & 152.48 & 412.52 & 36.04 & 3290.21 \\
Node2Vec (walk\_length=5)   & \textbf{3.64} & \textbf{3.84} & 154.83 & 417.46 & \textbf{35.44} & 3217.85 \\
FUSE    & 12.67 & 13.30 & \textbf{49.15} & \textbf{84.88} & 96.76 & \textbf{1360.30} \\
\midrule
\multicolumn{7}{c}{\textbf{30-70 Split}} \\
\midrule
DeepWalk (walk\_length=5)    & 3.72 & 3.81 & 155.53 & 421.54 & 36.01 & 3143.32 \\
Node2Vec (walk\_length=5)    & \textbf{3.65} & \textbf{3.80} & 154.27 & 423.19 & \textbf{35.90} & 3141.81 \\
FUSE    & 14.25 & 14.55 & \textbf{64.63} & \textbf{111.87} & 104.89 & 1\textbf{698.52} \\
\bottomrule
\end{tabular}
\caption{Runtime comparison (in seconds) of DeepWalk (walk\_length=5), Node2Vec (walk\_length=5) and FUSE across datasets under 70-30 and 30-70 train-test splits for a fixed seed. The least runtimes have been highlited in \textbf{bold}.}
\label{tab:runtimes_deepwalk_node2vec}
\end{table}

\subsection{Experiments on different masking mechanisms} \label{app:missingness}
We also performed experiments on various masking rates and mechanisms to investigate the robustness of our method. We analysed our method on 3 types of simulated masking mechanisms, based on the 3 types of missingness as described in~\cite{missingness}. The notations of MCAR, MAR and MNAR have been redefined for our specific use case. We describe these mechanisms here:
\begin{itemize}
    \item Masking-Completely-At-Random (MCAR): The probability of a node label being masked is independent of the data.
    \item Masking-At-Random (MAR): The probability of a node label being masked is dependent on the feature vector of the node.
    \item Masking-Not-At-Random (MNAR): The probability of a node label being masked depends both on the feature vector of the node, and the label itself.
\end{itemize}
We simulated these masking scenarios using a procedure similar to~\cite{HyperImpute}, where the masks were generated using a logistic model with random coefficients. Further details can be found in the attached code. For each masking scenario, we tested 3 masking rates: $0.2, 0.5$ and $0.8$, and reported the mean and standard deviations of the classification accuracy and F1 score over 10 iterations with different random seeds. The MultiLayer Perceptron (MLP) was chosen to have depth and width equivalent to the graph neural network models, in this case 2 and 16 respectively. The associated results are given in Tables~\ref{tab:maskingMCARCora}--~\ref{tab:maskingMNARCiteseer}.

\begin{table}[ht]
    \centering
    \small
    \renewcommand{\arraystretch}{1.2}
    \begin{tabular}{l|l|ccc}
        \hline
        \textbf{Classifier} & \textbf{Embedding} & \textbf{Accuracy (\%) } & \textbf{F1 Score} & \textbf{Time (s)} \\
        \hline
        \multirow{9}{*}{\textbf{GCN}} & Random & 22.58 & 0.05 & 0.4303 \\
          & DeepWalk & \underline{51.43} & \textbf{0.27} & 12996.76 \\
          & Node2Vec & 50.32 & \underline{0.25} & 12038.33 \\
          & VGAE & 16.17 & 0.01 & 1098.25 \\
          & DGI & 16.16 & 0.01 & 758.04 \\
          & FUSE & \textbf{59.65} & \underline{0.25} & 1698.52 \\
          & GraFN & 26.28 & 0.08 & \textbf{360.64} \\
          & ReVAR & 16.14 & 0.01 & \underline{468.55} \\
          & Given & 41.22 & 0.10 & 0.0521 \\
        \hline
        \multirow{9}{*}{\textbf{GAT}} & Random & 19.08 & 0.02 & 0.4303 \\
          & DeepWalk & \textbf{67.65} & \textbf{0.44} & 12996.76\\
          & Node2Vec & \underline{66.86} & \textbf{0.44} & 12038.33\\
          & VGAE & 16.16 & 0.01 & \underline{1098.25} \\
          & DGI & 16.16 & 0.01 & \textbf{758.04} \\
          & FUSE & 63.83 & \underline{0.43} & 1698.52 \\
          & GraFN & 57.04 & 0.39 & 13712.21 \\
          & ReVAR & 16.16 & 0.01 & 9265.35 \\
          & Given & 56.74 & 0.28 & 0.0521 \\
        \hline
        \multirow{9}{*}{\textbf{SAGE}} & Random & 15.13 & 0.02 & 0.4303 \\
          & DeepWalk & \textbf{61.95} & 0.23 & 12996.76 \\
          & Node2Vec & \underline{61.75} & \underline{0.24} & 12038.33 \\
          & VGAE & 16.16 & 0.01 & 1098.25 \\
          & DGI & 16.16 & 0.01 & 758.04 \\
          & FUSE & 59.65 & \textbf{0.25} & 1698.52\\
          & GraFN & 36.49 & 0.13 & \textbf{199.11} \\
          & ReVAR & 15.81 & 0.01 & \underline{285.55} \\
          & Given & 53.65 & 0.18 & 0.0521 \\
        \hline
    \end{tabular}
    \caption{Performance of different embedding–classifier pairs (except GraFN and ReVAR as they do have degenerate embedders and classifiers) on the ArXiv dataset (30–70 split) for a fixed seed. Embedding generation times were added across each of the embeddings except GraFN and ReVAR for which the time required by each encoder is given separately. The best and second-best in each metric, for each classifier are highlighted in \textbf{bold} and \underline{underline} respectively.}
    \label{tab:arxiv_embeddings_30_70}
\end{table}

\begin{table}[ht]
\centering
\small
\renewcommand{\arraystretch}{1.2}
\setlength{\tabcolsep}{6pt}
\begin{tabular}{l|l|ccc}
\hline
\textbf{Classifier} & \textbf{Embedding} & \textbf{Accuracy} & \textbf{F1 Score} & \textbf{Time (s)} \\
\hline
\multirow{9}{*}{\textbf{GCN}} & Random & 33.19 & 0.4628 & \\
 & DeepWalk & \textbf{50.04} & \underline{0.2180} & 13029.78\\
 & Node2Vec & 49.20 & 0.1992 & 12899.23\\
 & VGAE & 13.12 & 0.0058 & 1072.42\\
 & DGI & 16.37 & 0.0070 & 633.06\\
 & FUSE & \underline{49.97} & \textbf{0.2353} & 1360.30\\
 & GraFN & 26.21 & 0.07 & \textbf{360.17} \\
 & ReVAR & 16.37 & 0.01 & \underline{432.88} \\
 & Given & 38.19 & 0.0794 & 0.0473 \\
\hline
\multirow{9}{*}{\textbf{GAT}} & Random & 22.37 & 0.0300 & \underline{0.4628} \\
 & DeepWalk & \textbf{68.58} & \underline{0.4601} & 13029.78 \\
 & Node2Vec & \underline{67.87} & 0.4506 & 12899.23 \\
 & VGAE & 13.44 & 0.0082 & \underline{1072.42} \\
 & DGI & 13.13 & 0.0073 & \textbf{633.06} \\
 & FUSE & 67.45 & \textbf{0.4682} & 1360.30 \\
 & GraFN & 61.35 & 0.43 & 13564.56 \\
 & ReVAR & 16.37 & 0.01 & 9462.59 \\
 & Given & 58.74 & 0.3294 & 0.0473 \\
\hline
\multirow{9}{*}{\textbf{SAGE}} & Random & 16.15 & 0.0163 & 0.4628 \\
 & DeepWalk & \textbf{62.39} & \textbf{0.2421} & 13029.78 \\
 & Node2Vec & \underline{62.03} & \textbf{0.2421} & 12899.73 \\
 & VGAE & 16.53 & 0.0092 & 1072.42 \\
 & DGI & 16.37 & 0.0070 & 633.06\\
 & FUSE & 61.91 & \underline{0.2344} & 1360.30\\
 & GraFN & 44.73 & 0.17 & \textbf{248.13} \\
 & ReVAR & 16.13 & 0.01 & \underline{321.53} \\
 & Given & 54.16 & 0.1792 & 0.0473 \\
\hline
\end{tabular}
\caption{Performance of different embedding–classifier pairs (except GraFN and ReVAR as they have degenerate embedders and classifiers) on the ArXiv dataset (70–30 split) for a fixed seed. Embedding generation times were added across each of the embeddings except GraFN and ReVAR for which the time required by each encoder is given separately. The best and second-best in each metric, for each classifier are highlighted in \textbf{bold} and \underline{underline} respectively.}
\label{tab:arxiv_embeddings_70_30}
\end{table}

\begin{table}[ht]
    \centering
    \scriptsize
    \begin{tabular}{l|ccc|ccc}
\toprule
 & & \textbf{Accuracy} & & & \textbf{F1 Score} & \\
 \textbf{Rates} & 0.2 & 0.5 & 0.8 & 0.2 & 0.5 & 0.8 \\
\midrule
\multicolumn{7}{c}{\textbf{GCN}}\\
\midrule
FUSE & \textbf{0.81} $\pm 0.02$ & \textbf{0.78} $\pm 0.01$ & \textbf{0.77} $\pm 0.01$ & \textbf{0.80} $\pm 0.02$ & \textbf{0.77} $\pm 0.01$ & \textbf{0.76} $\pm 0.01$ \\
Node2Vec & \underline{0.78} $\pm 0.01$ & \underline{0.76} $\pm 0.01$ & \underline{0.68} $\pm 0.04$ & \underline{0.76} $\pm 0.02$ & \underline{0.75} $\pm 0.01$ & \underline{0.67} $\pm 0.04$ \\
DeepWalk & \underline{0.78} $\pm 0.01$ & 0.75 $\pm 0.02$ & 0.71 $\pm 0.01$ & \underline{0.76} $\pm 0.01$ & 0.74 $\pm 0.02$ & .69 $\pm 0.01$ \\
VGAE & \underline{0.78} $\pm 0.01$ & 0.74 $\pm 0.02$ & 0.66 $\pm 0.02$ & \underline{0.76} $\pm 0.01$ & 0.72 $\pm 0.02$ & 0.65 $\pm 0.02$ \\
DGI & 0.32 $\pm 0.05$ & 0.36 $\pm 0.07$ & 0.32 $\pm 0.05$ & 0.10 $\pm 0.08$ & 0.16 $\pm 0.12$ & 0.15 $\pm 0.10$ \\
Random & 0.52 $\pm 0.02$ & 0.39 $\pm 0.02$ & 0.29 $\pm 0.03$ & 0.50 $\pm 0.02$ & 0.35 $\pm 0.02$ & 0.25 $\pm 0.03$ \\
\midrule
\multicolumn{7}{c}{\textbf{GAT}}\\
\midrule
FUSE & \textbf{0.84} $\pm 0.02$ & \textbf{0.82} $\pm 0.01$ & \textbf{0.77} $\pm 0.02$ & \textbf{0.83} $\pm 0.02$ & \textbf{0.81} $\pm 0.01$ & \textbf{0.75} $\pm 0.02$ \\
Node2Vec & \underline{0.83} $\pm 0.01$ & \underline{0.80} $\pm 0.01$ & \underline{0.74} $\pm 0.02$ & \underline{0.82} $\pm 0.02$ & \underline{0.79} $\pm 0.01$ & \underline{0.73} $\pm 0.02$ \\
DeepWalk & \textbf{0.84} $\pm 0.02$ & \underline{0.80} $\pm 0.02$ & \underline{0.74} $\pm 0.02$ & \textbf{0.83} $\pm 0.02$ & \underline{0.78} $\pm 0.02$ & \underline{0.73} $\pm 0.02$ \\
VGAE & 0.79 $\pm 0.02$ & 0.75 $\pm 0.02$ & 0.71 $\pm 0.02$ & 0.78 $\pm 0.02$ & 0.73 $\pm 0.02$ & 0.69 $\pm 0.02$ \\
DGI & 0.70 $\pm 0.05$ & 0.64 $\pm 0.08$ & 0.60 $\pm 0.07$ & 0.68 $\pm 0.08$ & 0.59 $\pm 0.10$ & 0.56 $\pm 0.09$ \\
Random & 0.66 $\pm 0.02$ & 0.50 $\pm 0.03$ & 0.33 $\pm 0.04$ & 0.63 $\pm 0.03$ & 0.47 $\pm 0.03$ & 0.27 $\pm 0.04$ \\
\midrule
\multicolumn{7}{c}{\textbf{SAGE}}\\
\midrule
FUSE & \underline{0.85} $\pm 0.02$ & \underline{0.82} $\pm 0.01$ & \underline{0.76} $\pm 0.01$ & \textbf{0.84} $\pm 0.02$ & \underline{0.81} $\pm 0.01$ & \underline{0.74} $\pm 0.01$ \\
Node2Vec & \underline{0.85} $\pm 0.02$ & \textbf{0.83} $\pm 0.01$ & \underline{0.77} $\pm 0.01$ & \textbf{0.84} $\pm 0.02$ & \underline{0.81} $\pm 0.01$ & \textbf{0.76} $\pm 0.01$ \\
DeepWalk & \textbf{0.86} $\pm 0.01$ & \textbf{0.83} $\pm 0.01$ & \textbf{0.78} $\pm 0.01$ & \textbf{0.84} $\pm 0.01$ & \textbf{0.82} $\pm 0.01$ & \textbf{0.76} $\pm 0.01$ \\
VGAE & 0.79 $\pm 0.02$ & 0.73 $\pm 0.01$ & 0.67 $\pm 0.02$ & \underline{0.78} $\pm 0.02$ & 0.71 $\pm 0.01$ & 0.64 $\pm 0.02$ \\
DGI & 0.60 $\pm 0.05$ & 0.59 $\pm 0.04$ & 0.58 $\pm 0.04$ & 0.52 $\pm 0.09$ & 0.50 $\pm 0.08$ & 0.50 $\pm 0.07$ \\
Random & 0.51 $\pm 0.02$ & 0.35 $\pm 0.02$ & 0.26 $\pm 0.02$ & 0.46 $\pm 0.03$ & 0.26 $\pm 0.02$ & 0.17 $\pm 0.02$ \\
\midrule
\multicolumn{7}{c}{\textbf{MLP}}\\
\midrule
FUSE & 0.81 $\pm 0.02$ & 0.79 $\pm 0.01$ & 0.73 $\pm 0.01$ & 0.79 $\pm 0.03$ & 0.77 $\pm 0.01$ & 0.71 $\pm 0.01$ \\
Node2Vec & \textbf{0.84} $\pm 0.01$ & \textbf{0.82} $\pm 0.01$ & \textbf{0.76} $\pm 0.01$ & \underline{0.83} $\pm 0.01$ & \textbf{0.81} $\pm 0.01$ & \underline{0.74} $\pm 0.02$ \\
DeepWalk & \underline{0.85} $\pm 0.01$ & \underline{0.81} $\pm 0.01$ & \underline{0.77} $\pm 0.02$ & \textbf{0.84} $\pm 0.01$ & \underline{0.80} $\pm 0.01$ & \textbf{0.75} $\pm 0.02$ \\
VGAE & 0.65 $\pm 0.02$ & 0.63 $\pm 0.02$ & 0.62 $\pm 0.01$ & 0.63 $\pm 0.02$ & 0.61 $\pm 0.01$ & 0.60 $\pm 0.02$ \\
DGI & 0.53 $\pm 0.05$ & 0.49 $\pm 0.07$ & 0.48 $\pm 0.06$ & 0.44 $\pm 0.09$ & 0.35 $\pm 0.13$ & 0.36 $\pm 0.11$ \\
Random & 0.18 $\pm 0.02$ & 0.18 $\pm 0.01$ & 0.19 $\pm 0.01$ & 0.15 $\pm 0.02$ & 0.14 $\pm 0.01$ & 0.15 $\pm 0.01$ \\
\bottomrule
\end{tabular}
\caption{Classification experiments on different masking rates for the MCAR scenario on the Cora dataset. The mean and standard deviation over 10 iterations is reported. The best and second-best in each metric, for each masking rate and each classifier are highlighted in \textbf{bold} and \underline{underline} respectively.}
    \label{tab:maskingMCARCora}
\end{table}

\begin{table}[ht]
    \centering
    \scriptsize
    \begin{tabular}{l|ccc|ccc}
\toprule
  & & \textbf{Accuracy} & & & \textbf{F1 Score} & \\
  \textbf{Rates} & 0.2 & 0.5 & 0.8 & 0.2 & 0.5 & 0.8 \\
\midrule
 \multicolumn{7}{c}{\textbf{GCN}}\\
 \hline
 FUSE & $\bm{0.81} \pm 0.01$ & $\bm{0.78} \pm 0.01$ & $\bm{0.76} \pm 0.02$ & $\bm{0.80} \pm 0.01$ & $\bm{0.76} \pm 0.01$ & $\bm{0.75} \pm 0.02$ \\
 Node2Vec & $\underline{0.79} \pm 0.02$ & $0.76 \pm 0.01$ & $\underline{0.68} \pm 0.02$ & $\underline{0.77} \pm 0.02$ & $\underline{0.75} \pm 0.02$ & $\underline{0.66} \pm 0.02$ \\
 DeepWalk & $0.77 \pm 0.02$ & $\underline{0.77} \pm 0.02$ & $\underline{0.68} \pm 0.02$ & $0.76 \pm 0.02$ & $\bm{0.76} \pm 0.02$ & $\underline{0.66} \pm 0.02$ \\
 VGAE & $0.77 \pm 0.02$ & $0.72 \pm 0.02$ & $0.66 \pm 0.03$ & $0.76 \pm 0.02$ & $0.72 \pm 0.02$ & $0.64 \pm 0.03$ \\
 DGI & $0.28 \pm 0.02$ & $0.30 \pm 0.03$ & $0.36 \pm 0.08$ & $0.06 \pm 0.00$ & $0.08 \pm 0.04$ & $0.20 \pm 0.14$ \\
 Random & $0.51 \pm 0.02$ & $0.40 \pm 0.02$ & $0.29 \pm 0.03$ & $0.48 \pm 0.03$ & $0.36 \pm 0.02$ & $0.24 \pm 0.03$ \\
\midrule
 \multicolumn{7}{c}{\textbf{GAT}}\\
 \midrule
 FUSE & $\bm{0.85} \pm 0.01$ & $\bm{0.81} \pm 0.01$ & $\bm{0.77} \pm 0.01$ & $\bm{0.85} \pm 0.01$ & $\bm{0.80} \pm 0.01$ & $\bm{0.76} \pm 0.02$ \\
 Node2Vec & $\underline{0.84} \pm 0.01$ & $\underline{0.80} \pm 0.01$ & $\underline{0.75} \pm 0.02$ & $\underline{0.83} \pm 0.01$ & $\underline{0.79} \pm 0.01$ & $0.73 \pm 0.02$ \\
 DeepWalk & $0.83 \pm 0.01$ & $\underline{0.80} \pm 0.01$ & $\underline{0.75} \pm 0.01$ & $0.82 \pm 0.01$ & $\underline{0.79} \pm 0.01$ & $\underline{0.74} \pm 0.01$ \\
 VGAE & $0.78 \pm 0.01$ & $0.75 \pm 0.01$ & $0.70 \pm 0.01$ & $0.77 \pm 0.01$ & $0.73 \pm 0.01$ & $0.68 \pm 0.02$ \\
 DGI & $0.68 \pm 0.05$ & $0.68 \pm 0.03$ & $0.64 \pm 0.03$ & $0.64 \pm 0.08$ & $0.67 \pm 0.03$ & $0.62 \pm 0.04$ \\
 Random & $0.65 \pm 0.03$ & $0.50 \pm 0.03$ & $0.34 \pm 0.03$ & $0.63 \pm 0.03$ & $0.46 \pm 0.04$ & $0.25 \pm 0.04$ \\
\midrule
\multicolumn{7}{c}{\textbf{SAGE}}\\
\midrule
 FUSE & $\bm{0.85} \pm 0.01$ & $\underline{0.81} \pm 0.01$ & $\underline{0.76} \pm 0.02$ & $\bm{0.84} \pm 0.01$ & $\underline{0.80} \pm 0.01$ & $\underline{0.75} \pm 0.02$ \\
 Node2Vec & $\bm{0.85} \pm 0.01$ & $\bm{0.83} \pm 0.01$ & $\bm{0.78} \pm 0.01$ & $\bm{0.84} \pm 0.02$ & $\bm{0.82} \pm 0.01$ & $\bm{0.77} \pm 0.01$ \\
 DeepWalk & $\bm{0.85} \pm 0.02$ & $\bm{0.83} \pm 0.01$ & $\bm{0.78} \pm 0.01$ & $\underline{0.83} \pm 0.02$ & $\bm{0.82} \pm 0.01$ & $\bm{0.77} \pm 0.02$ \\
 VGAE & $\underline{0.76} \pm 0.02$ & $0.72 \pm 0.01$ & $0.67 \pm 0.01$ & $0.74 \pm 0.02$ & $0.70 \pm 0.01$ & $0.63 \pm 0.03$ \\
 DGI & $0.57 \pm 0.07$ & $0.59 \pm 0.04$ & $0.53 \pm 0.06$ & $0.47 \pm 0.08$ & $0.51 \pm 0.06$ & $0.42 \pm 0.10$ \\
 Random & $0.49 \pm 0.02$ & $0.35 \pm 0.02$ & $0.27 \pm 0.01$ & $0.43 \pm 0.03$ & $0.27 \pm 0.03$ & $0.17 \pm 0.01$ \\
\midrule
\multicolumn{7}{c}{\textbf{MLP}}\\
\midrule
 FUSE & $\underline{0.80} \pm 0.02$ & $0.77 \pm 0.01$ & $\underline{0.72} \pm 0.02$ & $0.79 \pm 0.02$ & $\underline{0.76} \pm 0.01$ & $\underline{0.70} \pm 0.02$ \\
 Node2Vec & $\bm{0.84} \pm 0.01$ & $\underline{0.81} \pm 0.01$ & $\bm{0.76} \pm 0.01$ & $\bm{0.83} \pm 0.01$ & $\bm{0.81} \pm 0.01$ & $\bm{0.75} \pm 0.01$ \\
 DeepWalk & $\bm{0.84} \pm 0.02$ & $\bm{0.82} \pm 0.01$ & $\bm{0.76} \pm 0.01$ & $\underline{0.82} \pm 0.02$ & $\bm{0.81} \pm 0.01$ & $\bm{0.75} \pm 0.01$ \\
 VGAE & $0.65 \pm 0.01$ & $0.63 \pm 0.01$ & $0.61 \pm 0.02$ & $0.63 \pm 0.02$ & $0.61 \pm 0.01$ & $0.59 \pm 0.02$ \\
 DGI & $0.54 \pm 0.03$ & $0.50 \pm 0.07$ & $0.50 \pm 0.06$ & $0.44 \pm 0.07$ & $0.40 \pm 0.12$ & $0.39 \pm 0.11$ \\
 Random & $0.17 \pm 0.01$ & $0.18 \pm 0.01$ & $0.19 \pm 0.01$ & $0.14 \pm 0.01$ & $0.14 \pm 0.01$ & $0.14 \pm 0.01$ \\
\bottomrule
\end{tabular}
    \caption{Classification experiments on different masking rates for the MAR scenario on the Cora dataset. The mean and standard deviation over 10 iterations is reported. The best and second-best in each metric, for each masking rate and each classifier are highlighted in \textbf{bold} and \underline{underline} respectively.}
    \label{tab:maskingMARCora}
\end{table}

\begin{table}[ht]
    \centering
    \scriptsize
    \begin{tabular}{l|ccc|ccc}
\toprule
  & & \textbf{Accuracy} & & & \textbf{F1 Score} & \\
  \textbf{Rates} & 0.2 & 0.5 & 0.8 & 0.2 & 0.5 & 0.8 \\
\midrule
 \multicolumn{7}{c}{\textbf{GCN}}\\
 \midrule
 FUSE & $\bm{0.80} \pm 0.01$ & $\bm{0.78} \pm 0.02$ & $\bm{0.76} \pm 0.02$ & $\bm{0.79} \pm 0.01$ & $\bm{0.76} \pm 0.01$ & $\bm{0.74} \pm 0.02$ \\
 Node2Vec & $0.76 \pm 0.05$ & $\underline{0.75} \pm 0.02$ & $0.66 \pm 0.02$ & $0.74 \pm 0.06$ & $0.73 \pm 0.02$ & $0.63 \pm 0.02$ \\
 DeepWalk & $\underline{0.78} \pm 0.02$ & $\underline{0.75} \pm 0.03$ & $\underline{0.68} \pm 0.03$ & $\underline{0.76} \pm 0.03$ & $\underline{0.74} \pm 0.03$ & $\underline{0.65} \pm 0.04$ \\
 VGAE & $0.77 \pm 0.02$ & $0.73 \pm 0.01$ & $0.64 \pm 0.03$ & $0.75 \pm 0.02$ & $0.72 \pm 0.01$ & $0.61 \pm 0.03$ \\
 DGI & $0.30 \pm 0.03$ & $0.32 \pm 0.05$ & $0.32 \pm 0.09$ & $0.08 \pm 0.03$ & $0.12 \pm 0.08$ & $0.16 \pm 0.11$ \\
 Random & $0.48 \pm 0.03$ & $0.40 \pm 0.03$ & $0.29 \pm 0.04$ & $0.45 \pm 0.03$ & $0.36 \pm 0.03$ & $0.24 \pm 0.03$ \\
\midrule
 \multicolumn{7}{c}{\textbf{GAT}}\\
 \midrule
 FUSE & $\underline{0.84} \pm 0.02$ & $\bm{0.80} \pm 0.01$ & $\bm{0.75} \pm 0.02$ & $\bm{0.83} \pm 0.02$ & $\underline{0.78} \pm 0.01$ & $\bm{0.74} \pm 0.02$ \\
 Node2Vec & $\underline{0.84} \pm 0.02$ & $\bm{0.80} \pm 0.02$ & $0.73 \pm 0.02$ & $\bm{0.83} \pm 0.02$ & $\bm{0.79} \pm 0.02$ & $0.71 \pm 0.02$ \\
 DeepWalk & $\bm{0.85} \pm 0.01$ & $\bm{0.80} \pm 0.02$ & $\underline{0.74} \pm 0.02$ & $\bm{0.83} \pm 0.02$ & $\bm{0.79} \pm 0.02$ & $\underline{0.72} \pm 0.02$ \\
 VGAE & $0.77 \pm 0.02$ & $\underline{0.73} \pm 0.01$ & $0.69 \pm 0.01$ & $\underline{0.75} \pm 0.02$ & $0.72 \pm 0.01$ & $0.68 \pm 0.02$ \\
 DGI & $0.61 \pm 0.10$ & $0.68 \pm 0.04$ & $0.59 \pm 0.06$ & $0.58 \pm 0.13$ & $0.65 \pm 0.06$ & $0.54 \pm 0.07$ \\
 Random & $0.64 \pm 0.02$ & $0.49 \pm 0.03$ & $0.32 \pm 0.04$ & $0.62 \pm 0.03$ & $0.44 \pm 0.04$ & $0.24 \pm 0.04$ \\
\midrule
\multicolumn{7}{c}{\textbf{SAGE}}\\
\midrule
 FUSE & $\underline{0.85} \pm 0.01$ & $\underline{0.80} \pm 0.02$ & $0.74 \pm 0.02$ & $\underline{0.83} \pm 0.02$ & $0.79 \pm 0.02$ & $0.72 \pm 0.02$ \\
 Node2Vec & $\bm{0.86} \pm 0.01$ & $\bm{0.83} \pm 0.01$ & $\underline{0.76} \pm 0.02$ & $\bm{0.85} \pm 0.01$ & $\bm{0.82} \pm 0.01$ & $\underline{0.73} \pm 0.03$ \\
 DeepWalk & $\underline{0.85} \pm 0.01$ & $\bm{0.83} \pm 0.01$ & $\bm{0.77} \pm 0.01$ & $\underline{0.83} \pm 0.01$ & $\underline{0.81} \pm 0.01$ & $\bm{0.75} \pm 0.02$ \\
 VGAE & $0.75 \pm 0.02$ & $0.71 \pm 0.02$ & $0.65 \pm 0.02$ & $0.73 \pm 0.02$ & $0.69 \pm 0.02$ & $0.61 \pm 0.03$ \\
 DGI & $0.55 \pm 0.05$ & $0.57 \pm 0.05$ & $0.53 \pm 0.05$ & $0.47 \pm 0.08$ & $0.48 \pm 0.08$ & $0.43 \pm 0.06$ \\
 Random & $0.49 \pm 0.03$ & $0.35 \pm 0.02$ & $0.25 \pm 0.02$ & $0.43 \pm 0.03$ & $0.25 \pm 0.03$ & $0.17 \pm 0.01$ \\
\midrule
\multicolumn{7}{c}{\textbf{MLP}}\\
\midrule
 FUSE & $\underline{0.81} \pm 0.01$ & $0.76 \pm 0.01$ & $0.71 \pm 0.02$ & $0.79 \pm 0.01$ & $0.74 \pm 0.02$ & $0.69 \pm 0.02$ \\
 Node2Vec & $\bm{0.85} \pm 0.01$ & $\bm{0.82} \pm 0.01$ & $\underline{0.75} \pm 0.01$ & $\bm{0.84} \pm 0.02$ & $\bm{0.81} \pm 0.01$ & $\underline{0.73} \pm 0.02$ \\
 DeepWalk & $\bm{0.85} \pm 0.01$ & $\underline{0.81} \pm 0.01$ & $\bm{0.76} \pm 0.02$ & $\underline{0.83} \pm 0.01$ & $\underline{0.80} \pm 0.01$ & $\bm{0.74} \pm 0.03$ \\
 VGAE & $0.63 \pm 0.02$ & $0.63 \pm 0.02$ & $0.60 \pm 0.01$ & $0.61 \pm 0.01$ & $0.61 \pm 0.03$ & $0.58 \pm 0.02$ \\
 DGI & $0.50 \pm 0.06$ & $0.48 \pm 0.09$ & $0.49 \pm 0.04$ & $0.41 \pm 0.09$ & $0.35 \pm 0.14$ & $0.39 \pm 0.06$ \\
 Random & $0.18 \pm 0.01$ & $0.18 \pm 0.01$ & $0.18 \pm 0.01$ & $0.14 \pm 0.02$ & $0.14 \pm 0.01$ & $0.14 \pm 0.01$ \\
\bottomrule
\end{tabular}
    \caption{Classification experiments on different masking rates for the MNAR scenario on the Cora dataset. The mean and standard deviation over 10 iterations is reported. The best and second-best in each metric, for each masking rate and each classifier are highlighted in \textbf{bold} and \underline{underline} respectively.}
    \label{tab:maskingMNARCora}
\end{table}

\begin{table}
    \centering
    \scriptsize
    \begin{tabular}{l|ccc|ccc}
\toprule
  & & \textbf{Accuracy} & & & \textbf{F1 Score} & \\
  \textbf{Rates} & 0.2 & 0.5 & 0.8 & 0.2 & 0.5 & 0.8 \\
\midrule
 \multicolumn{7}{c}{\textbf{GCN}}\\
 \midrule
 FUSE & $\bm{0.66} \pm 0.01$ & $\bm{0.67} \pm 0.01$ & $\bm{0.59} \pm 0.01$ & $\bm{0.63} \pm 0.01$ & $\bm{0.64} \pm 0.01$ & $\bm{0.55} \pm 0.01$ \\
 Node2Vec & $\underline{0.58} \pm 0.02$ & $\underline{0.54} \pm 0.01$ & $\underline{0.46} \pm 0.02$ & $\underline{0.52} \pm 0.02$ & $\underline{0.50} \pm 0.01$ & $0\underline{.42} \pm 0.02$ \\
 DeepWalk & $0.57 \pm 0.02$ & $0.53 \pm 0.01$ & $0.44 \pm 0.01$ & $\underline{0.52} \pm 0.02$ & $\underline{0.50} \pm 0.01$ & $0.41 \pm 0.01$ \\
 VGAE & $0.54 \pm 0.02$ & $0.50 \pm 0.01$ & $0.42 \pm 0.02$ & $0.50 \pm 0.02$ & $0.46 \pm 0.01$ & $0.38 \pm 0.02$ \\
 DGI & $0.30 \pm 0.07$ & $0.32 \pm 0.06$ & $0.32 \pm 0.03$ & $0.19 \pm 0.09$ & $0.20 \pm 0.09$ & $0.25 \pm 0.04$ \\
 Random & $0.34 \pm 0.03$ & $0.28 \pm 0.02$ & $0.24 \pm 0.02$ & $0.32 \pm 0.03$ & $0.26 \pm 0.02$ & $0.21 \pm 0.02$ \\
\midrule
 \multicolumn{7}{c}{\textbf{GAT}}\\
 \midrule
 FUSE & $\bm{0.72} \pm 0.01$ & $\bm{0.68} \pm 0.01$ & $\bm{0.59} \pm 0.01$ & $\underline{0.68} \pm 0.01$ & $\bm{0.64} \pm 0.01$ & $\bm{0.55} \pm 0.01$ \\
 Node2Vec & $\underline{0.71} \pm 0.02$ & $\underline{0.65} \pm 0.01$ & $\underline{0.56} \pm 0.01$ & $\bm{0.69} \pm 0.02$ & $\underline{0.62} \pm 0.01$ & $\underline{0.53} \pm 0.01$ \\
 DeepWalk & $\underline{0.71} \pm 0.01$ & $0.64 \pm 0.01$ & $0.55 \pm 0.02$ & $0.67 \pm 0.01$ & $0.61 \pm 0.01$ & $0.52 \pm 0.01$ \\
 VGAE & $0.61 \pm 0.02$ & $0.56 \pm 0.02$ & $0.47 \pm 0.02$ & $0.57 \pm 0.02$ & $0.52 \pm 0.01$ & $0.43 \pm 0.02$ \\
 DGI & $0.49 \pm 0.03$ & $0.48 \pm 0.02$ & $0.45 \pm 0.02$ & $0.42 \pm 0.05$ & $0.43 \pm 0.02$ & $0.40 \pm 0.02$ \\
 Random & $0.48 \pm 0.02$ & $0.40 \pm 0.01$ & $0.28 \pm 0.02$ & $0.45 \pm 0.02$ & $0.37 \pm 0.01$ & $0.25 \pm 0.01$ \\
\midrule
\multicolumn{7}{c}{\textbf{SAGE}}\\
\midrule
 FUSE & $\bm{0.72} \pm 0.01$ & $\bm{0.67} \pm 0.01$ & $\bm{0.58} \pm 0.01$ & $\bm{0.69} \pm 0.01$ & $\bm{0.63} \pm 0.01$ & $\bm{0.54} \pm 0.01$ \\
 Node2Vec & $0.70 \pm 0.01$ & $\underline{0.66} \pm 0.01$ & $\underline{0.57} \pm 0.01$ & $0.66 \pm 0.01$ & $\underline{0.62} \pm 0.01$ & $\bm{0.54} \pm 0.02$ \\
 DeepWalk & $\underline{0.71} \pm 0.01$ & $\underline{0.66} \pm 0.01$ & $\underline{0.57} \pm 0.01$ & $\underline{0.67} \pm 0.01$ & $\underline{0.62} \pm 0.01$ & $\bm{0.54} \pm 0.01$ \\
 VGAE & $0.57 \pm 0.02$ & $0.50 \pm 0.01$ & $0.44 \pm 0.02$ & $0.51 \pm 0.02$ & $0.46 \pm 0.01$ & $\underline{0.40} \pm 0.01$ \\
 DGI & $0.45 \pm 0.03$ & $0.46 \pm 0.02$ & $0.42 \pm 0.01$ & $0.38 \pm 0.03$ & $0.40 \pm 0.02$ & $0.35 \pm 0.02$ \\
 Random & $0.38 \pm 0.03$ & $0.29 \pm 0.01$ & $0.22 \pm 0.01$ & $0.33 \pm 0.02$ & $0.25 \pm 0.01$ & $0.19 \pm 0.01$ \\
\midrule
\multicolumn{7}{c}{\textbf{MLP}}\\
\midrule
 FUSE & $\bm{0.72} \pm 0.01$ & $\bm{0.66} \pm 0.01$ & $\bm{0.57} \pm 0.01$ & $0.67 \pm 0.01$ & $\underline{0.62} \pm 0.01$ & $\bm{0.53} \pm 0.01$ \\
 Node2Vec & $\bm{0.72} \pm 0.02$ & $\underline{0.65} \pm 0.01$ & $\underline{0.56} \pm 0.01$ & $\bm{0.69} \pm 0.02$ & $\underline{0.62} \pm 0.01$ & $\bm{0.53} \pm 0.02$ \\
 DeepWalk & $\underline{0.71} \pm 0.01$ & $\bm{0.66} \pm 0.01$ & $0.55 \pm 0.01$ & $\underline{0.68} \pm 0.02$ & $\bm{0.63} \pm 0.01$ & $\underline{0.52} \pm 0.01$ \\
 VGAE & $0.42 \pm 0.02$ & $0.40 \pm 0.01$ & $0.38 \pm 0.01$ & $0.38 \pm 0.01$ & $0.37 \pm 0.01$ & $0.36 \pm 0.01$ \\
 DGI & $0.39 \pm 0.07$ & $0.41 \pm 0.03$ & $0.37 \pm 0.04$ & $0.31 \pm 0.09$ & $0.34 \pm 0.05$ & $0.30 \pm 0.05$ \\
 Random & $0.18 \pm 0.01$ & $0.17 \pm 0.01$ & $0.17 \pm 0.01$ & $0.17 \pm 0.01$ & $0.16 \pm 0.01$ & $0.16 \pm 0.01$ \\
\bottomrule
\end{tabular}
    \caption{Classification experiments on different masking rates for the MCAR scenario on the CiteSeer dataset. The mean and standard deviation over 10 iterations is reported. The best and second-best in each metric, for each masking rate and each classifier are highlighted in \textbf{bold} and \underline{underline} respectively.}
    \label{tab:maskingMCARCiteseer}
\end{table}

\begin{table}
    \centering
    \scriptsize
    \begin{tabular}{l|ccc|ccc}
\toprule
  & & \textbf{Accuracy} & & & \textbf{F1 Score} & \\
  \textbf{Rates} & 0.2 & 0.5 & 0.8 & 0.2 & 0.5 & 0.8 \\
\midrule
 \multicolumn{7}{c}{\textbf{GCN}}\\
 \midrule
 FUSE & $\bm{0.68} \pm 0.01$ & $\bm{0.68} \pm 0.01$ & $\bm{0.58} \pm 0.01$ & $\bm{0.64} \pm 0.01$ & $\bm{0.64} \pm 0.01$ & $\bm{0.55} \pm 0.01$ \\
 Node2Vec & $0.57 \pm 0.01$ & $\underline{0.54} \pm 0.01$ & $\underline{0.42} \pm 0.02$ & $0.51 \pm 0.02$ & $\underline{0.51} \pm 0.01$ & $0.39 \pm 0.02$ \\
 DeepWalk & $\underline{0.58} \pm 0.02$ & $\underline{0.54} \pm 0.02$ & $\underline{0.42} \pm 0.03$ & $\underline{0.52} \pm 0.02$ & $\underline{0.51} \pm 0.02$ & $\underline{0.40} \pm 0.02$ \\
 VGAE & $0.54 \pm 0.03$ & $0.48 \pm 0.02$ & $0.41 \pm 0.03$ & $0.51 \pm 0.03$ & $0.45 \pm 0.02$ & $0.38 \pm 0.03$ \\
 DGI & $0.29 \pm 0.10$ & $0.33 \pm 0.07$ & $0.32 \pm 0.02$ & $0.18 \pm 0.13$ & $0.21 \pm 0.10$ & $0.23 \pm 0.05$ \\
 Random & $0.34 \pm 0.02$ & $0.26 \pm 0.01$ & $0.23 \pm 0.02$ & $0.32 \pm 0.01$ & $0.24 \pm 0.01$ & $0.21 \pm 0.02$ \\
\midrule
 \multicolumn{7}{c}{\textbf{GAT}}\\
 \midrule
 FUSE & $\bm{0.72} \pm 0.01$ & $\bm{0.68} \pm 0.01$ & $\bm{0.59} \pm 0.01$ & $\bm{0.68} \pm 0.01$ & $\bm{0.64} \pm 0.01$ & $\bm{0.54} \pm 0.01$ \\
 Node2Vec & $\underline{0.71} \pm 0.02$ & $\underline{0.65} \pm 0.02$ & $\underline{0.54} \pm 0.03$ & $\underline{0.67} \pm 0.02$ & $\underline{0.61} \pm 0.01$ & $\underline{0.51} \pm 0.02$ \\
 DeepWalk & $\underline{0.71} \pm 0.01$ & $\underline{0.65} \pm 0.02$ & $\underline{0.54} \pm 0.02$ & $\underline{0.67} \pm 0.01$ & $\underline{0.61} \pm 0.02$ & $\underline{0.51} \pm 0.02$ \\
 VGAE & $0.62 \pm 0.01$ & $0.57 \pm 0.01$ & $0.47 \pm 0.01$ & $0.58 \pm 0.02$ & $0.54 \pm 0.01$ & $0.43 \pm 0.01$ \\
 DGI & $0.51 \pm 0.02$ & $0.48 \pm 0.03$ & $0.44 \pm 0.03$ & $0.45 \pm 0.02$ & $0.42 \pm 0.04$ & $0.39 \pm 0.03$ \\
 Random & $0.48 \pm 0.02$ & $0.38 \pm 0.02$ & $0.27 \pm 0.02$ & $0.44 \pm 0.02$ & $0.35 \pm 0.02$ & $0.25 \pm 0.02$ \\
\midrule
\multicolumn{7}{c}{\textbf{SAGE}}\\
\midrule
 FUSE & $\bm{0.72} \pm 0.01$ & $\bm{0.67} \pm 0.01$ & $\bm{0.58} \pm 0.01$ & $\bm{0.68} \pm 0.01$ & $\bm{0.63} \pm 0.01$ & $\bm{0.54} \pm 0.01$ \\
 Node2Vec & $\underline{0.71} \pm 0.01$ & $\underline{0.66} \pm 0.01$ & $\underline{0.57} \pm 0.02$ & $\underline{0.66} \pm 0.02$ & $\underline{0.62} \pm 0.01$ & $\bm{0.54} \pm 0.01$ \\
 DeepWalk & $\underline{0.71} \pm 0.01$ & $\underline{0.66} \pm 0.01$ & $\underline{0.57} \pm 0.01$ & $0\underline{.66} \pm 0.01$ & $\underline{0.62} \pm 0.01$ & $\underline{0.53} \pm 0.01$ \\
 VGAE & $0.57 \pm 0.01$ & $0.51 \pm 0.02$ & $0.43 \pm 0.01$ & $0.52 \pm 0.01$ & $0.47 \pm 0.02$ & $0.39 \pm 0.02$ \\
 DGI & $0.47 \pm 0.03$ & $0.45 \pm 0.03$ & $0.42 \pm 0.03$ & $0.40 \pm 0.02$ & $0.38 \pm 0.03$ & $0.35 \pm 0.04$ \\
 Random & $0.36 \pm 0.02$ & $0.29 \pm 0.02$ & $0.21 \pm 0.01$ & $0.31 \pm 0.02$ & $0.25 \pm 0.01$ & $0.19 \pm 0.01$ \\
\midrule
\multicolumn{7}{c}{\textbf{MLP}}\\
\midrule
 FUSE & $\underline{0.71} \pm 0.01$ & $\bm{0.66} \pm 0.01$ & $\bm{0.57} \pm 0.01$ & $\underline{0.66} \pm 0.01$ & $\underline{0.62} \pm 0.01$ & $\bm{0.53} \pm 0.01$ \\
 Node2Vec & $\bm{0.72} \pm 0.01$ & $\bm{0.66} \pm 0.01$ & $0.55 \pm 0.02$ & $\bm{0.68} \pm 0.01$ & $\underline{0.62} \pm 0.01$ & $\underline{0.52} \pm 0.01$ \\
 DeepWalk & $\bm{0.72} \pm 0.01$ & $\bm{0.66} \pm 0.02$ & $\underline{0.56} \pm 0.01$ & $\bm{0.68} \pm 0.01$ & $\bm{0.63} \pm 0.02$ & $\bm{0.53} \pm 0.01$ \\
 VGAE & $0.42 \pm 0.02$ & $\underline{0.41} \pm 0.01$ & $0.39 \pm 0.01$ & $0.40 \pm 0.02$ & $0.38 \pm 0.01$ & $0.36 \pm 0.01$ \\
 DGI & $0.42 \pm 0.05$ & $\underline{0.41} \pm 0.02$ & $0.37 \pm 0.03$ & $0.35 \pm 0.05$ & $0.34 \pm 0.03$ & $0.30 \pm 0.05$ \\
 Random & $0.17 \pm 0.01$ & $0.18 \pm 0.01$ & $0.18 \pm 0.01$ & $0.16 \pm 0.01$ & $0.16 \pm 0.01$ & $0.17 \pm 0.00$ \\
\bottomrule
\end{tabular}
    \caption{Classification experiments on different masking rates for the MAR scenario on the CiteSeer dataset. The mean and standard deviation over 10 iterations is reported. The best and second-best in each metric, for each masking rate and each classifier are highlighted in \textbf{bold} and \underline{underline} respectively.}
    \label{tab:maskingMARCiteseer}
\end{table}

\begin{table}
    \centering
    \scriptsize
    \begin{tabular}{l|ccc|ccc}
\toprule
  & & \textbf{Accuracy} & & & \textbf{F1 Score} & \\
  \textbf{Rates} & 0.2 & 0.5 & 0.8 & 0.2 & 0.5 & 0.8 \\
\midrule
 \multicolumn{7}{c}{\textbf{GCN}}\\
 \midrule
 FUSE & $\bm{0.68} \pm 0.02$ & $\bm{0.68} \pm 0.01$ & $\bm{0.58} \pm 0.01$ & $\bm{0.64} \pm 0.02$ & $\bm{0.64} \pm 0.01$ & $\bm{0.54} \pm 0.01$ \\
 Node2Vec & $0.58 \pm 0.02$ & $\underline{0.53} \pm 0.01$ & $0.42 \pm 0.02$ & $0.52 \pm 0.02$ & $\underline{0.50} \pm 0.01$ & $0.39 \pm 0.02$ \\
 DeepWalk & $\underline{0.59} \pm 0.02$ & $\underline{0.53} \pm 0.01$ & $\underline{0.43} \pm 0.02$ & $\underline{0.54} \pm 0.02$ & $0.49 \pm 0.01$ & $\underline{0.40} \pm 0.02$ \\
 VGAE & $0.56 \pm 0.01$ & $0.49 \pm 0.02$ & $0.38 \pm 0.03$ & $0.51 \pm 0.02$ & $0.46 \pm 0.01$ & $0.35 \pm 0.02$ \\
 DGI & $0.31 \pm 0.10$ & $0.32 \pm 0.05$ & $0.28 \pm 0.05$ & $0.20 \pm 0.12$ & $0.23 \pm 0.07$ & $0.18 \pm 0.06$ \\
 SGCL & $0.36 \pm 0.01$ & $0.26 \pm 0.02$ & $0.21 \pm 0.02$ & $0.33 \pm 0.01$ & $0.24 \pm 0.02$ & $0.19 \pm 0.02$ \\
 Random & $0.35 \pm 0.01$ & $0.27 \pm 0.01$ & $0.21 \pm 0.03$ & $0.32 \pm 0.01$ & $0.25 \pm 0.01$ & $0.19 \pm 0.02$ \\
\midrule
 \multicolumn{7}{c}{\textbf{GAT}}\\
 \midrule
 FUSE & $\bm{0.73} \pm 0.02$ & $\bm{0.69} \pm 0.01$ & $\bm{0.58} \pm 0.01$ & $\bm{0.68} \pm 0.02$ & $\bm{0.64} \pm 0.01$ & $\bm{0.54} \pm 0.01$ \\
 Node2Vec & $\underline{0.71} \pm 0.01$ & $0\underline{.65} \pm 0.02$ & $\underline{0.55} \pm 0.02$ & $\underline{0.67} \pm 0.02$ & $\underline{0.61} \pm 0.01$ & $\underline{0.52} \pm 0.02$ \\
 DeepWalk & $\bm{0.73} \pm 0.02$ & $\underline{0.65} \pm 0.02$ & $\underline{0.55} \pm 0.03$ & $\underline{0.67} \pm 0.02$ & $\underline{0.61} \pm 0.02$ & $\underline{0.52} \pm 0.02$ \\
 VGAE & $0.62 \pm 0.01$ & $0.56 \pm 0.01$ & $0.44 \pm 0.02$ & $0.58 \pm 0.02$ & $0.52 \pm 0.01$ & $0.42 \pm 0.02$ \\
 DGI & $0.52 \pm 0.03$ & $0.48 \pm 0.04$ & $0.41 \pm 0.03$ & $0.44 \pm 0.03$ & $0.42 \pm 0.05$ & $0.36 \pm 0.03$ \\
 Random & $0.47 \pm 0.02$ & $0.38 \pm 0.01$ & $0.25 \pm 0.01$ & $0.44 \pm 0.02$ & $0.35 \pm 0.02$ & $0.22 \pm 0.01$ \\
\midrule
 \multicolumn{7}{c}{\textbf{SAGE}}\\
 \midrule
 FUSE & $\bm{0.73} \pm 0.02$ & $\bm{0.67} \pm 0.01$ & $\bm{0.57} \pm 0.01$ & $\bm{0.68} \pm 0.02$ & $\bm{0.63} \pm 0.01$ & $\underline{0.53} \pm 0.01$ \\
 Node2Vec & $0.70 \pm 0.01$ & $\underline{0.66} \pm 0.01$ & $\bm{0.57} \pm 0.01$ & $0.64 \pm 0.01$ & $\underline{0.62} \pm 0.01$ & $\underline{0.53} \pm 0.01$ \\
 DeepWalk & $\underline{0.72} \pm 0.02$ & $\bm{0.67} \pm 0.01$ & $\bm{0.57} \pm 0.01$ & $\underline{0.66} \pm 0.03$ & $\underline{0.62} \pm 0.01$ & $\bm{0.54} \pm 0.01$ \\
 VGAE & $0.57 \pm 0.02$ & $0.52 \pm 0.02$ & $\underline{0.42} \pm 0.02$ & $0.51 \pm 0.02$ & $0.47 \pm 0.02$ & $0.39 \pm 0.02$ \\
 DGI & $0.48 \pm 0.04$ & $0.47 \pm 0.02$ & $0.39 \pm 0.02$ & $0.40 \pm 0.03$ & $0.40 \pm 0.03$ & $0.33 \pm 0.04$ \\
 Random & $0.37 \pm 0.03$ & $0.27 \pm 0.02$ & $0.21 \pm 0.01$ & $0.32 \pm 0.03$ & $0.24 \pm 0.02$ & $0.17 \pm 0.01$ \\
\midrule
 \multicolumn{7}{c}{\textbf{MLP}}\\
 \midrule
 FUSE & $\underline{0.72} \pm 0.02$ & $\bm{0.66} \pm 0.02$ & $\underline{0.55} \pm 0.01$ & $\underline{0.67} \pm 0.02$ & $\underline{0.62} \pm 0.02$ & $\underline{0.52} \pm 0.01$ \\
 Node2Vec & $0.71 \pm 0.01$ & $\bm{0.66} \pm 0.01$ & $\bm{0.56}\pm 0.02$ & $\underline{0.67} \pm 0.01$ & $\underline{0.62} \pm 0.01$ & $\bm{0.53} \pm 0.02$ \\
 DeepWalk & $\bm{0.73} \pm 0.01$ & $\bm{0.66} \pm 0.01$ & $\underline{0.55} \pm 0.02$ & $\bm{0.68} \pm 0.02$ & $\bm{0.63} \pm 0.02$ & $\underline{0.52} \pm 0.01$ \\
 VGAE & $0.42 \pm 0.01$ & $\underline{0.40} \pm 0.01$ & $0.37 \pm 0.01$ & $0.38 \pm 0.01$ & $0.38 \pm 0.01$ & $0.35 \pm 0.01$ \\
 DGI & $0.41 \pm 0.04$ & $\underline{0.40} \pm 0.03$ & $0.37 \pm 0.03$ & $0.33 \pm 0.04$ & $0.34 \pm 0.03$ & $0.31 \pm 0.03$ \\
 Random & $0.18 \pm 0.02$ & $0.18 \pm 0.01$ & $0.17 \pm 0.01$ & $0.16 \pm 0.02$ & $0.16 \pm 0.01$ & $0.16 \pm 0.01$ \\
\bottomrule
\end{tabular}
    \caption{Classification experiments on different masking rates for the MNAR scenario on the CiteSeer dataset. The mean and standard deviation over 10 iterations is reported. The best and second-best in each metric, for each masking rate and each classifier are highlighted in \textbf{bold} and \underline{underline} respectively.}
    \label{tab:maskingMNARCiteseer}
\end{table}

\end{document}